\documentclass{article}
\PassOptionsToPackage{table}{xcolor}
\usepackage{iclr2027_conference,times}

\usepackage{amsmath,amsfonts,bm}

\def\eqref#1{equation~\ref{#1}}

\def\1{\bm{1}}

\DeclareMathAlphabet{\mathsfit}{\encodingdefault}{\sfdefault}{m}{sl}
\SetMathAlphabet{\mathsfit}{bold}{\encodingdefault}{\sfdefault}{bx}{n}

\makeatletter
\DeclareRobustCommand\onedot{\futurelet\@let@token\@onedot}
\def\@onedot{\ifx\@let@token.\else.\null\fi\xspace}
\def\eg{\emph{e.g}\onedot,~} 
\def\ie{\emph{i.e}\onedot,~}

\makeatother

\usepackage[utf8]{inputenc}
\usepackage[T1]{fontenc}
\usepackage{url}
\usepackage{booktabs}
\usepackage{amsmath}
\usepackage{amssymb}
\usepackage{amsfonts}
\usepackage{nicefrac}
\usepackage{microtype}
\usepackage{subcaption}
\usepackage{xcolor}
\usepackage{colortbl}
\definecolor{refblue}{rgb}{0.21,0.49,0.74}
\usepackage[pagebackref,breaklinks,colorlinks,allcolors=refblue]{hyperref}
\usepackage{graphicx}
\usepackage{xspace}
\usepackage{caption}
\usepackage{multirow}
\usepackage{enumitem}
\usepackage{etoc}
\usepackage{algorithm}
\usepackage{algpseudocode}
\usepackage[capitalize]{cleveref}
\crefname{section}{Section}{Sections}
\crefname{table}{Table}{Tables}
\crefname{figure}{Figure}{Figures}
\crefname{equation}{Equation}{Equations}
\crefname{appendix}{Appendix}{Appendices}
\definecolor{oursteal}{HTML}{50EBCE}
\definecolor{stereored}{HTML}{D62728}
\definecolor{hawgray}{HTML}{4C4C4C}
\definecolor{bcpresent}{HTML}{1B9E77}
\definecolor{bcheld}{HTML}{D95F02}
\definecolor{bcabsent}{HTML}{9A9A9A}
\newcommand{\best}[1]{\cellcolor{oursteal!60}\textbf{#1}}
\newcommand{\second}[1]{\cellcolor{oursteal!18}#1}
\newcommand{\bestcap}[1]{{\setlength{\fboxsep}{1pt}\colorbox{oursteal!60}{\textbf{#1}}}}
\newcommand{\secondcap}[1]{{\setlength{\fboxsep}{1pt}\colorbox{oursteal!18}{#1}}}

\title{RLHND: Video Foundation Models as \\ Physically Grounded Hand Trackers \\ for Robot Learning}

\author{%
Seungjun Moon$^{1,2}$ \quad Subin Jeon$^{3}$ \quad Sangwoo Kim$^{1}$ \quad Hanbyul Joo$^{3}$ \quad Jinwoo Shin$^{1,2}$ \\
$^{1}$RLWRLD \quad $^{2}$KAIST \quad $^{3}$Seoul National University \\
\texttt{seungjun.moon@rlwrld.ai}
}

\iclrfinalcopy
\renewcommand{\iclrheadertext}{Preprint}
\begin{document}

\maketitle

\etocdepthtag.toc{maintext}
\begin{abstract}
Recently, approaches that leverage human video datasets for robot policy training have become increasingly prevalent.
However, most existing hand trackers regress pose from cropped frames with limited priors on hand motion and object interaction, resulting in inaccurate and physically inconsistent estimates.
Moreover, the lack of physical cues, \eg contact and force, limits the use of human videos for robot policy training.
To this end, we propose RLHND, a video foundation model-based hand tracking model that jointly estimates hand pose and realistic tactile information from monocular egocentric videos.
RLHND turns the pre-trained Cosmos~3 video diffusion backbone into a deterministic clip-level feature extractor via clean-latent conditioning, carrying its learned priors on hand motion and hand-object interaction into tracking.
For pose estimation, RLHND (i) predicts hand poses with anatomically plausible joint angles and (ii) enables optional conditioning on the shape parameter to maintain consistent hand shape within the same video and even across videos recorded by the same actor.
For tactile estimation, a separate tactile expert stream, trained with the pose stream frozen, predicts dense contact and force over the hand surface.
We further adopt LBS-based feature spreading to enable vertex-wise feature extraction without costly per-vertex attention.
RLHND achieves state-of-the-art performance across various benchmark datasets for pose estimation, while also achieving state-of-the-art performance in contact and force estimation.
Moreover, we demonstrate the utility of RLHND for robot learning through retargeting results and real-world robot experiments.
The code will be publicly available at \url{https://seungjun-moon.github.io/rlhnd/}.
 \end{abstract}

\section{Introduction}
Along with the rapid advancement of vision-language models~\citep{openai2023gpt4, liu2023visual, team2023gemini, bai2025qwen25vl}, robotics foundation models have also made significant progress in recent years~\citep{brohan2023rt2, team2024octo, kim2024openvla, black2024pi0, intelligence2025pi05, bjorck2025gr00t, kim2026rldx1}.
However, robotics datasets remain substantially more limited in scale than vision-language datasets and are considerably more labor-intensive to collect.
Recently, researchers have recognized the potential of human egocentric videos, which are relatively easy to collect and can capture a wide range of tasks that are difficult to demonstrate through teleoperation.
Consequently, human datasets~\citep{grauman2022ego4d, banerjee2024introducing, hoque2026egodex} have been increasingly leveraged to train robotics foundation models~\citep{ye2025lapa, kareer2024egomimic, yang2025egovla, luo2025beingh0, kim2026rldx1}.

However, utilizing human videos for robot training requires bridging two major gaps between human videos and real-world robot episodes.
The first is the \emph{action-space discrepancy} between humans and robots.
Existing works address this discrepancy by retargeting human hand motion to a specific robot embodiment~\citep{handa2020dexpilot,qin2023anyteleop,shaw2022videodex,lepert2025phantom}, defining a co-action space that encompasses both human and robot joint actions~\citep{kareer2024egomimic,luo2025beingh0,li2025vitra}, or adopting an end-effector-based action space consisting of wrist poses and keypoints~\citep{haldar2025pointpolicy,liu2025egozero,kim2026dexpointpolicy,cai2025innon}.

\begin{figure}[t]
\centering
\includegraphics[width=\linewidth]{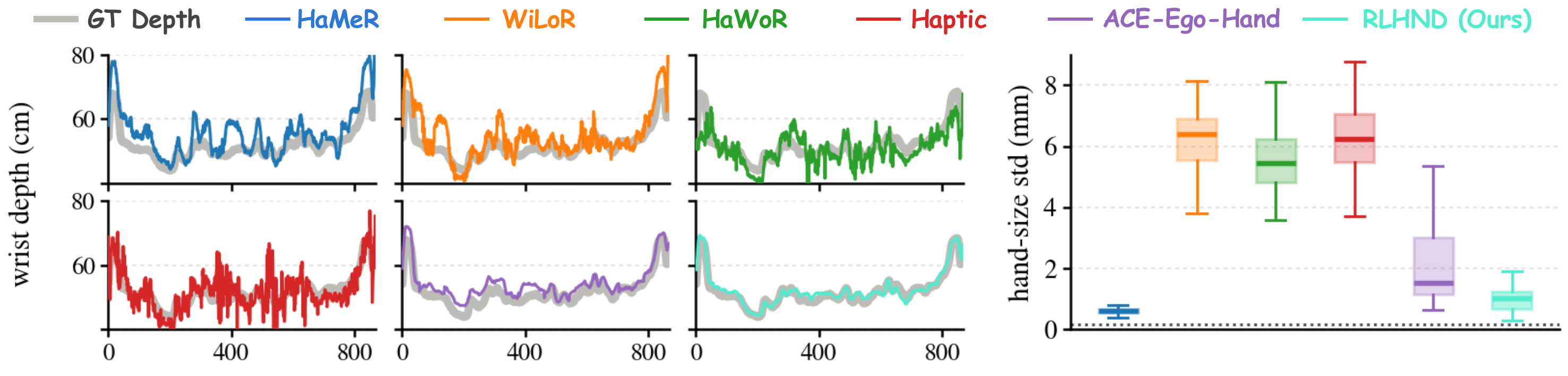}
\caption{
\textbf{3D instability of hand motion reconstruction methods.}
\emph{Left:} Wrist depth over 850 frames from an ARCTIC test-set clip.
Baselines exhibit severe oscillation or drift relative to the ground-truth depth.
\emph{Right:} Per-video standard deviation of the estimated hand size.
Even within a single video, existing hand motion estimators exhibit $\sim$8 mm variation in estimated hand size.
}
\vspace{-0.5cm}
\label{fig:motivation}
\end{figure}

Regardless of the choice of action space, these approaches require human action labels that provide 3D hand joint locations.
However, existing hand motion reconstruction methods are often optimized primarily for 2D reprojection or image-space alignment, which can make their 3D predictions unreliable for robot action labeling.
As shown in Figure~\ref{fig:motivation}, existing hand reconstructions exhibit severe jitter along the depth axis (left), while the estimated size of the same hand can fluctuate substantially across frames (right).
The latter arises from the inherent scale-depth ambiguity under perspective projection, where depth errors can be absorbed into the estimated hand shape.
Such temporally inconsistent and physically implausible 3D trajectories are problematic for robot learning, where small errors in motion estimation can lead to substantially different or infeasible actions.

The second gap is \emph{missing physical information}.
Recent vision-language-action models increasingly incorporate tactile input alongside vision~\citep{zhang2026human, niu2026trex, chen2026univtac, zhang2026unitacvla}, as tactile information can distinguish a successful grasp from merely hovering over an object.
However, human datasets generally cannot provide tactile information, except for those collected with pressure boards~\citep{grady2022pressurevision} or wearable tactile gloves~\citep{song2025opentouch}.
To this end, several vision-based contact and force estimation models~\citep{jeon2026hope, jung2026learning} have emerged, but these remain underexplored compared to hand pose estimation.

In this paper, we present RLHND, which produces both metrically accurate, anatomically realistic 3D bimanual hand motion and dense contact and force estimates over the hand surface from a monocular human video.
Since recent works~\citep{wang2026surprising, liu2026aceegohand} show that pre-trained video diffusion models internalize strong priors on hand-object interactions, RLHND uses Cosmos~3 Nano~\citep{nvidia2026cosmos3} as its video feature encoder.
Unlike \citet{liu2026aceegohand}, which feeds clean videos directly into a denoising backbone, we instead route them through the conditioning-frame interface of Cosmos~3, which is designed to receive clean context frames during pre-training.
This design keeps our fine-tuning aligned with the pre-trained interface of the backbone.

On top of the Cosmos~3 encoder, we make three design choices motivated by real-world deployment.
\emph{(i) Shape caching.} RLHND caches the shape parameter and uses a single value throughout the entire video.
This not only eliminates the hand-size fluctuations shown in Figure~\ref{fig:motivation}, but also allows direct injection of a pre-calibrated $\boldsymbol\beta$, which is particularly useful in real-world data collection settings where the hand shape of the operator is known.
\emph{(ii) Anatomical pose parameterization.} Rather than regressing the $45$ rotational DoF of MANO, we align each joint with its anatomical axes and freeze the axes about which a human finger cannot rotate, regressing $29$ DoF.
\emph{(iii) A decoupled tactile expert.} Contact and force are predicted by a second expert stream that is trained with the pose stream frozen, so tactile supervision never perturbs the pose.
To obtain vertex-wise features, we introduce LBS-based feature spreading, which uses the linear blend skinning weights of MANO to spread bone-level features to the $778$ MANO vertices, avoiding the need for costly vertex-wise attention.

For motion reconstruction, we evaluate RLHND on HOT3D and ARCTIC, as well as EgoDex, which is held out during training.
RLHND consistently outperforms all baselines across the three benchmark datasets.
For tactile estimation, RLHND achieves the best performance in both contact and force estimation among various benchmarks.
Finally, we demonstrate the utility of RLHND for robot learning by retargeting its human demonstrations to five dexterous hands and training Dexterous Point Policy~\citep{kim2026dexpointpolicy}.
RLHND produces more accurate and less jittery joint actions than the baselines, and training with keypoints from RLHND improves Dexterous Point Policy notably.

\textbf{Contributions.} We highlight the main contributions of our work as follows:
\begin{itemize}[leftmargin=1.5em,itemsep=2pt]
\item We extend video foundation model-based hand tracking beyond pose estimation: RLHND leverages the clean-latent conditioning interface of Cosmos~3 to extract rich spatiotemporal representations from monocular video and jointly estimates metrically accurate bimanual motion with dense contact and force over the hand surface.

\item We achieve state-of-the-art performance in both hand motion reconstruction and tactile estimation, enabled by physically grounded design choices including cached hand shape, anatomically constrained pose, and a decoupled tactile expert with LBS-based feature spreading.

\item We demonstrate the utility of RLHND for robot learning: its outputs yield accurate and smooth retargeted commands across five dexterous hands and enable training Dexterous Point Policy from human demonstrations with automatically labeled contact.
\end{itemize}
\section{Related Work}
\label{sec:related}

\paragraph{Hand motion reconstruction.}
Hand motion reconstruction methods have been developed to improve both pose accuracy and temporal consistency.
The primary approach uses image-based methods~\citep{pavlakos2024reconstructing, potamias2024wilor}, which regress MANO~\citep{romero2017embodied} parameters from a cropped image input by a hand detection model~\citep{potamias2024wilor}.
Since they treat every frame independently, they cannot consider any temporal consistency among frames.
In contrast, video-based methods~\citep{zhang2025hawor, ye2025haptic, xu2026handflow} add temporal context either by fusing crop features across frames~\citep{ye2025haptic} or by tracking the hand and the camera in a world frame~\citep{zhang2025hawor, xu2026handflow}.
Despite this, their reliance on per-frame crops causes them to inherit detector box jitter and remain only weakly consistent over time.
Most recently, models built on video foundation backbones take the entire clip as input and infer over both hands jointly~\citep{wang2026surprising,liu2026aceegohand}, which notably improves temporal coherence.
Even these models, however, regress a hand shape per window, leading to temporal variation in the estimated hand size, while the scale-depth ambiguity of perspective projection allows depth errors to manifest as variations in the estimated shape.
RLHND keeps the clip-level formulation but fixes a single shape per video and constrains the pose to the anatomical degrees of freedom, which removes both failure modes.

\paragraph{Hand contact and force estimation.}
Unlike motion reconstruction, force labels require specialized hardware, \eg pressure boards~\citep{grady2022pressurevision} or tactile gloves~\citep{song2025opentouch}.
PressureVisionDB~\citep{grady2022pressurevision} records RGB video of a bare hand pressing a planar capacitive pad, providing dense and finely resolved pressure maps that are spatially limited to the palm-pad contact region.
OpenTouch~\citep{song2025opentouch} instead moves the sensor onto the hand, pairing in-the-wild egocentric video with a $16{\times}16$ taxel grid on the palmar side of a wearable glove, enabling force measurements during natural manipulation of diverse objects.
EgoPressure~\citep{zhao2025egopressure} extends pressure sensing to egocentric video with multi-view MANO fits, while EgoTactile~\citep{zeng2026egotactile} provides full-hand glove pressure on $63$ everyday objects and a bare-hand subset.
However, differences in sensor calibration and hand-surface correspondence make it difficult to pool these datasets without converting their labels to a common representation.
For contact estimation, tactile hardware is not required: HACO~\citep{jung2026learning} generates hand contact labels from the distances between hand and object meshes in HOI datasets and uses them to train contact estimation models.
For force estimation, however, such geometric supervision is insufficient, requiring datasets with physical force measurements.
HOPE~\citep{jeon2026hope} unifies contact and force labels from pressure boards and tactile gloves on the MANO surface, enabling joint learning of contact and force estimation.
RLHND follows the unified label format of HOPE while introducing a decoupled tactile expert, achieving improved tactile estimation performance.
\section{Method}

\begin{figure}[t]
\centering
\includegraphics[width=\linewidth]{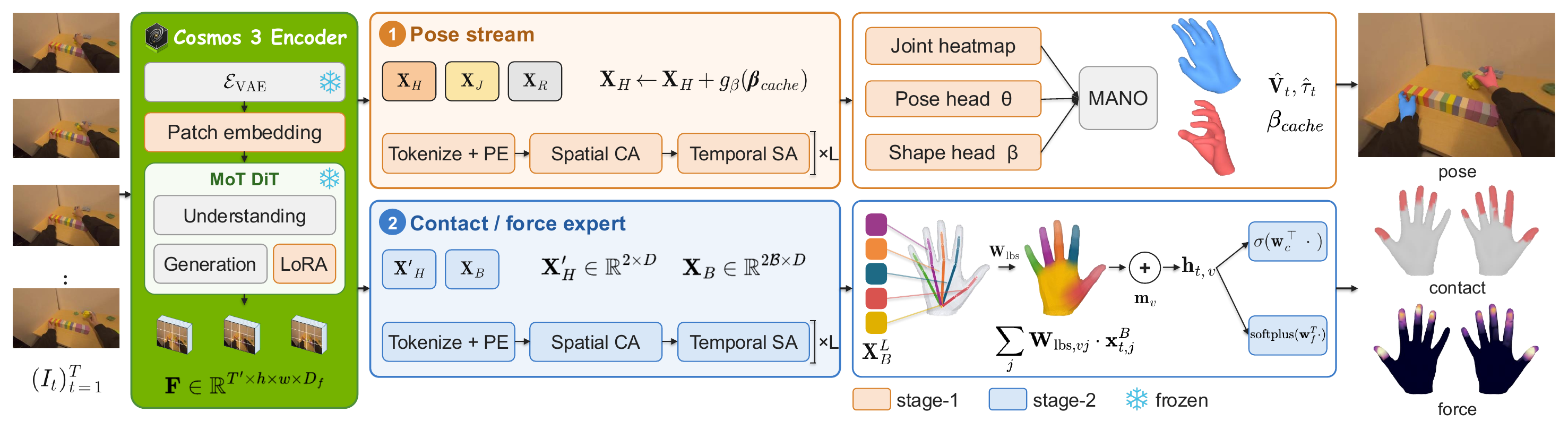}
\caption{\textbf{RLHND overview.}
A pre-trained Cosmos 3 backbone receives the clip through its conditioning-frame interface and is adapted with a trainable patch embedding and LoRA adapters.
The pose expert stream, trained in stage-1, yields MANO parameters and a metric camera-space translation
while the tactile expert stream is trained in stage-2 with the pose stream frozen.
}
\label{fig:structure}
\end{figure}

Given a monocular video, RLHND estimates both hand pose and tactile signals.
The video is processed as a sequence of $W$ clips $\mathcal{C}_{1},\dots,\mathcal{C}_{W}$, each holding $T$ consecutive frames $\mathcal{C}_{w} = (\mathbf{I}_{t})_{t=1}^{T}$.
The pose output consists of MANO~\citep{romero2017embodied} parameters, \ie shape $\hat{\boldsymbol\beta}^{w} \in \mathbb{R}^{10}$ and anatomically constrained articulation $\hat{\boldsymbol\theta}_{t}$ with the global wrist rotation, along with metric camera-space translation $\hat{\boldsymbol\tau}_{t} \in \mathbb{R}^{3}$ and per-frame hand presence and visibility.
The tactile output consists of per-vertex tactile signals: contact probability $\hat{\mathbf{c}}_{t} \in [0,1]^{|V|}$ and force $\hat{\mathbf{f}}_{t} \in \mathbb{R}_{\geq 0}^{|V|}$, where $|V| = 778$ is the number of MANO vertices.
The RLHND architecture consists of three parts:
the \emph{clean-latent encoder} converts a pre-trained video-diffusion backbone into a deterministic video feature extractor,
the \emph{pose expert stream} predicts pose, shape, and translation in MANO parameter form from the video features,
and the \emph{tactile expert stream} estimates contact and force from the same video features with its own weights, trained with the pose stream frozen, thereby decoupling tactile prediction from pose prediction.
Figure~\ref{fig:structure} illustrates the overall process of the proposed method.

\subsection{Clean-Latent Video Diffusion Encoder}
\label{method:encoder}
Since video diffusion models are known to internalize strong priors over hand motion and object interactions~\citep{wang2026surprising}, we use the pre-trained encoder of Cosmos~3~\citep{nvidia2026cosmos3}.
Its video tokenizer, \ie Wan2.2 VAE~\citep{wan2025wan}, encodes each clip into a clean latent $\mathbf{z} = \mathcal{E}_{\mathrm{VAE}}(\mathcal{C}_{w})$.
We fine-tune the patch embedding and generation pathway with LoRA~\citep{hu2022lora}, yielding a deterministic spatiotemporal feature grid $\mathbf{F} \in \mathbb{R}^{T' \times h \times w \times D_f}$.

Our use of the pre-trained VAE differs from prior work~\citep{wang2026surprising, liu2026aceegohand}, which feeds clean videos to Wan2.2 through its denoising interface by setting the out-of-distribution noise level $\sigma=0$.
Instead, we use the conditioning interface of Cosmos~3 for clean context frames, keeping our input scheme aligned with pre-training.
We also feed an empty prompt, since the video stream is already conditioned on the clean frames, whereas \citet{liu2026aceegohand} uses a fixed caption.

\subsection{Pose Expert Stream}
\label{method:decoder}
Following ACE-Ego-Hand~\citep{liu2026aceegohand}, we tokenize the $\mathbf{F}$ with spatial positional and Fourier-encoded ray embeddings, and decode the resulting tokens with alternating spatial cross-attention and bidirectional temporal self-attention.
We depart from this design in two aspects: how the two hand tokens $\mathbf{X}_{H}$ are formed with the shape conditioning, and how hand articulation is parameterized.

\paragraph{Shape conditioning.}
We construct the two hand tokens $\mathbf{X}_{H}$ from the cached MANO shape parameter $\boldsymbol\beta_{\mathrm{cache}}$, which enforces consistent hand geometry throughout the video and can also be obtained from a one-time calibration of the actor's hand.
Unlike ACE-Ego-Hand~\citep{liu2026aceegohand}, which initializes its hand tokens from fixed learnable parameters, our model can directly incorporate this pre-measured shape.
Let $v^{w}$ denote the fraction of frames in $\mathcal{C}_{w}$ in which the hand is predicted present.
The hand tokens are built from a learnable query $\mathbf{q}_{H}$ and a zero-initialized MLP $g_{\beta}$:
\begin{equation}
  \mathbf{X}_{H}^{w} = \mathbf{q}_{H} + g_{\beta}\!\left(\boldsymbol\beta_{\mathrm{cache}}\right),
  \qquad
  \boldsymbol\beta_{\mathrm{cache}} =
  \begin{cases}
    \hat{\boldsymbol\beta}^{w^{*}}, & w^{*} = \min\{\,w' : v^{w'} > T_{v}\,\} \text{ exists},\\[2pt]
    \hat{\boldsymbol\beta}^{w}, & \text{otherwise},
  \end{cases}
  \label{eq:beta_cache}
\end{equation}
The shape estimated from the first sufficiently visible clip is shared across the video, and earlier clips are decoded again once it becomes available.
The zero initialization of $g_{\beta}$ gradually introduces the cached shape during training.
We use the ground-truth $\boldsymbol\beta$ with probability $0.5$ and disable the shape-head loss for these samples.
Since the public benchmarks do not provide pre-calibrated hand shapes, we estimate $\boldsymbol\beta$ from the first sufficiently visible clip and reuse it throughout each video.

\paragraph{Constrained pose parameterization.}
For more faithful tracking of the ground-truth hand articulation, we anatomically constrain the DoF of MANO~\citep{yang2021cpf}.
While MANO allows $3$-axis rotation at each of its $15$ joints, resulting in $45$ DoF, the human hand has only $21$ DoF.
We therefore align the rotation axes with the anatomical structure and disable infeasible axes.
Following the anatomy-aware kinematics of \citet{yang2021cpf}, the rotation $\text{R}_{j}\in SO(3)$ of joint $j$ is factorized about its anatomically defined twist, spread, and bend axes
$(\mathbf{a}^{\text{twist}}{j},\mathbf{a}^{\text{spread}}_{j},\mathbf{a}^{\text{bend}}_{j})$:
\begin{equation}
  \text{R}_{j} \;=\;
  \exp\!\big(\varphi^{\text{twist}}_{j}\,[\mathbf{a}^{\text{twist}}_{j}]_{\times}\big)\,
  \exp\!\big(\varphi^{\text{spread}}_{j}\,[\mathbf{a}^{\text{spread}}_{j}]_{\times}\big)\,
  \exp\!\big(\varphi^{\text{bend}}_{j}\,[\mathbf{a}^{\text{bend}}_{j}]_{\times}\big),
  \label{eq:axis_decomposition}
\end{equation}
where $[\cdot]_{\times}$ is the skew-symmetric operator.
Then, the feasible set $\mathcal{C}$ is defined as below:
\begin{equation}
  \mathcal{C} \;=\; \big\{\boldsymbol\Theta \;:\;
  \varphi^{\text{twist}}_{j}=\varphi^{\text{spread}}_{j}=0 \;\;\forall j\in\text{PIP}\cup\text{DIP} \big\}.
  \label{eq:dof_constraint}
\end{equation}
We leave the thumb unconstrained because its carpometacarpal joint is a saddle joint with non-orthogonal, non-fixed axes, for which the twist-spread-bend decomposition does not isolate an infeasible axis.
This leaves $29$ DoF after constraining the remaining joints.
To train the model with this representation, we obtain constrained $29$-DoF labels by fitting the existing $45$-DoF MANO labels to the feasible set $\mathcal{C}$ via inverse kinematics~\citep{zhou2020monocular, li2021hybrik}:
\begin{equation}
  \widehat{\boldsymbol\Theta}_{c} \;=\;
  \operatorname*{arg\,min}_{\boldsymbol\Theta_{c}\in\mathcal{C}}\;
  \big\|\text{J}(\boldsymbol\Theta_{c};\boldsymbol\beta)-\text{J}(\boldsymbol\Theta;\boldsymbol\beta)\big\|_{2}^{2},
  \label{eq:ik_objective}
\end{equation}
where $\text{J}(\cdot\,;\boldsymbol\beta)$ returns the forward-kinematics joint positions under the shape $\boldsymbol\beta$ carried by the label.
We solve Eq.~(\ref{eq:ik_objective}) in the anatomy-aligned Euler coordinates using damped Gauss-Newton steps:
\begin{equation}
  \boldsymbol\varphi \leftarrow \boldsymbol\varphi + \big(\mathbf{G}^{\top}\mathbf{G} + \lambda\mathbf{I}\big)^{-1}\mathbf{G}^{\top}\boldsymbol\epsilon,
  \label{eq:ik_step}
\end{equation}
where $\boldsymbol\varphi$ contains the optimized finger DoF, $\mathbf{G}$ is the Jacobian of the joint positions with respect to $\boldsymbol\varphi$, and $\boldsymbol\epsilon$ stacks the joint and fingertip residuals.
Further details are provided in Appendix~\ref{app:ik_design}.

\paragraph{Output Heads.} The readout heads decode the hand tokens into pose, shape, depth, and presence predictions.
Metric camera-space translation is then recovered using the ray-based mixed-PnP solver of ACE-Ego-Hand~\citep{liu2026aceegohand}.
We refer to Appendix~\ref{app:camera} for details.

\subsection{Tactile Expert Stream}
\label{method:tactile}
Tactile supervision is far scarcer than pose supervision, as pressure-labeled datasets often provide constrained or no pose labels~\citep{grady2022pressurevision,grady2024pressurevision,song2025opentouch}.
We therefore decouple tactile estimation from pose estimation, preventing such training from affecting pose prediction.

\paragraph{Bone tokens.}

Attending over all $|V|$ vertices per frame is computationally prohibitive, so we perform tactile estimation at the hand-skeleton level.
Let $\mathcal{B}$ denote the $|\mathcal{B}|=16$ joints of the MANO kinematic tree.
Each joint drives one bone of the skeleton through the linear-blend-skinning weights, so we attach one \emph{bone token} per joint and hand, yielding $\mathbf{X}_{B}\in\mathbb{R}^{2|\mathcal{B}|\times D}$ with $|\mathcal{B}|\ll|V|$. Each bone token represents the features of the vertices skinned to its bone.

\paragraph{LBS-based vertex readout.}
We keep temporal reasoning at the bone level and expand the features to vertices only at the output using the MANO Linear-Blend-Skinning (LBS) weight $\mathbf{W}_{\mathbf{lbs}}\in\mathbb{R}^{|V|\times|\mathcal{B}|}$, the fixed matrix from MANO that maps features to vertices.
Combined with a learned vertex embedding $\mathbf{m}_{v}$, the per-vertex tactile feature is computed as follows:
\begin{equation}
    \mathbf{h}_{t,v} \;=\; \sum_{j\in\mathcal{B}} \mathbf{W}_{\mathbf{lbs},vj}\, \big[\mathbf{X}^{L}_{B}\big]_{t,j} \;+\; \mathbf{m}_{v},
    \label{eq:lbs_spread}
\end{equation}
where $[\mathbf{X}^{L}_{B}]_{t,j}$ is the token of node $j$ at frame $t$ in $\mathbf{X}^{L}_{B}$, obtained by interpolation from $T'$ to $T$.

Two small MLP heads read the per-vertex contact logit and the force-distribution logit from $\mathbf{h}_{t,v}$, and the per-hand total force $\hat{F}_{t}$ is read from the expert's hand token through a softplus; the per-vertex force is the total distributed over the vertices, $\hat{f}_{t,v} = \hat{F}_{t}\,\mathrm{softmax}_{v}(\cdot)$.

\subsection{Training}
\label{method:training}
At stage-1 training, we only train the pose expert stream with the following:
\begin{equation}
  \mathcal{L}_{\mathrm{pose}} =
  \lambda_{\mathrm{rot}}\mathcal{L}_{\mathrm{rot}} +
  \lambda_{\beta}\mathcal{L}_{\beta} +
  \lambda_{\mathrm{3D}}\mathcal{L}_{\mathrm{3D}} +
  \lambda_{\mathrm{2D}}\mathcal{L}_{\mathrm{2D}} +
  \lambda_{\tau}\mathcal{L}_{\tau} +
  \lambda_{\mathrm{pres}}\mathcal{L}_{\mathrm{pres}} +
  \lambda_{\mathrm{tmp}}\mathcal{L}_{\mathrm{tmp}} +
  \lambda_{\mathrm{ray}}\mathcal{L}_{\mathrm{ray}}.
  \label{eq:stage1_loss}
\end{equation}
At stage-2 training, the pose expert stream is frozen, and only the tactile expert stream is trained:
\begin{equation}
\mathcal{L}_{\mathrm{tactile}} =
  \lambda_{c}\,\mathrm{BCE}_{w}\!\left(\hat{c}_{t,v},\, c_{t,v}\right)
  + \lambda_{F}\,\tfrac{1}{|V|}\Big|\textstyle\sum_{v}\hat{f}_{t,v} - \sum_{v} f_{t,v}\Big|
  + \lambda_{\pi}\, \mathrm{CE}\!\left(\hat{\mathbf{f}}_{t}/\textstyle\sum_{v}\hat{f}_{t,v},\; \mathbf{f}_{t}/\textstyle\sum_{v} f_{t,v}\right).
\label{eq:stage2_loss}
\end{equation}
Since contact vertices are far rarer than non-contact vertices, we use a positively re-weighted BCE, \ie $\mathrm{BCE}_{w}$.
The force is supervised through its per-hand total, normalized by the number of vertices, and through its distribution over the vertices, weighted by $\lambda_{F}$ and $\lambda_{\pi}$, respectively.
We elucidate the details of each loss term in Appendix~\ref{app:loss}.

\section{Experiments}
\label{sec:experiments}

\begin{figure}[t]
\centering
\includegraphics[width=\linewidth]{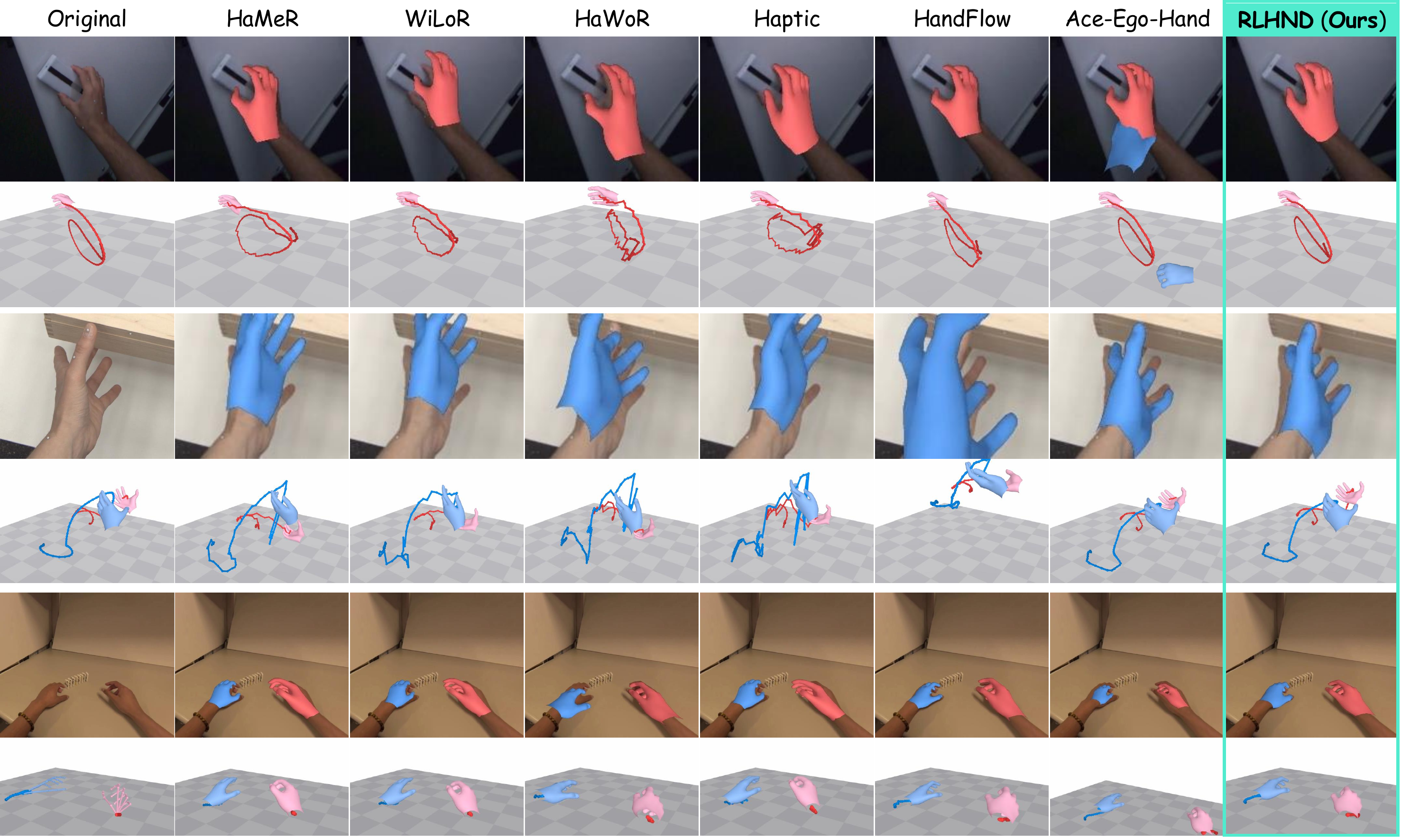}
\caption{
\textbf{Qualitative comparison of hand motion reconstruction.}
We visualize the reconstruction results from HOT3D (top), ARCTIC (middle), and EgoDex (bottom), with mesh overlays and trajectory visualizations.
Extensive visualization results can be found on the project page.
}
\label{fig:qualitative_pose}
\end{figure}

\begin{figure}[t]
\centering
\includegraphics[width=\linewidth]{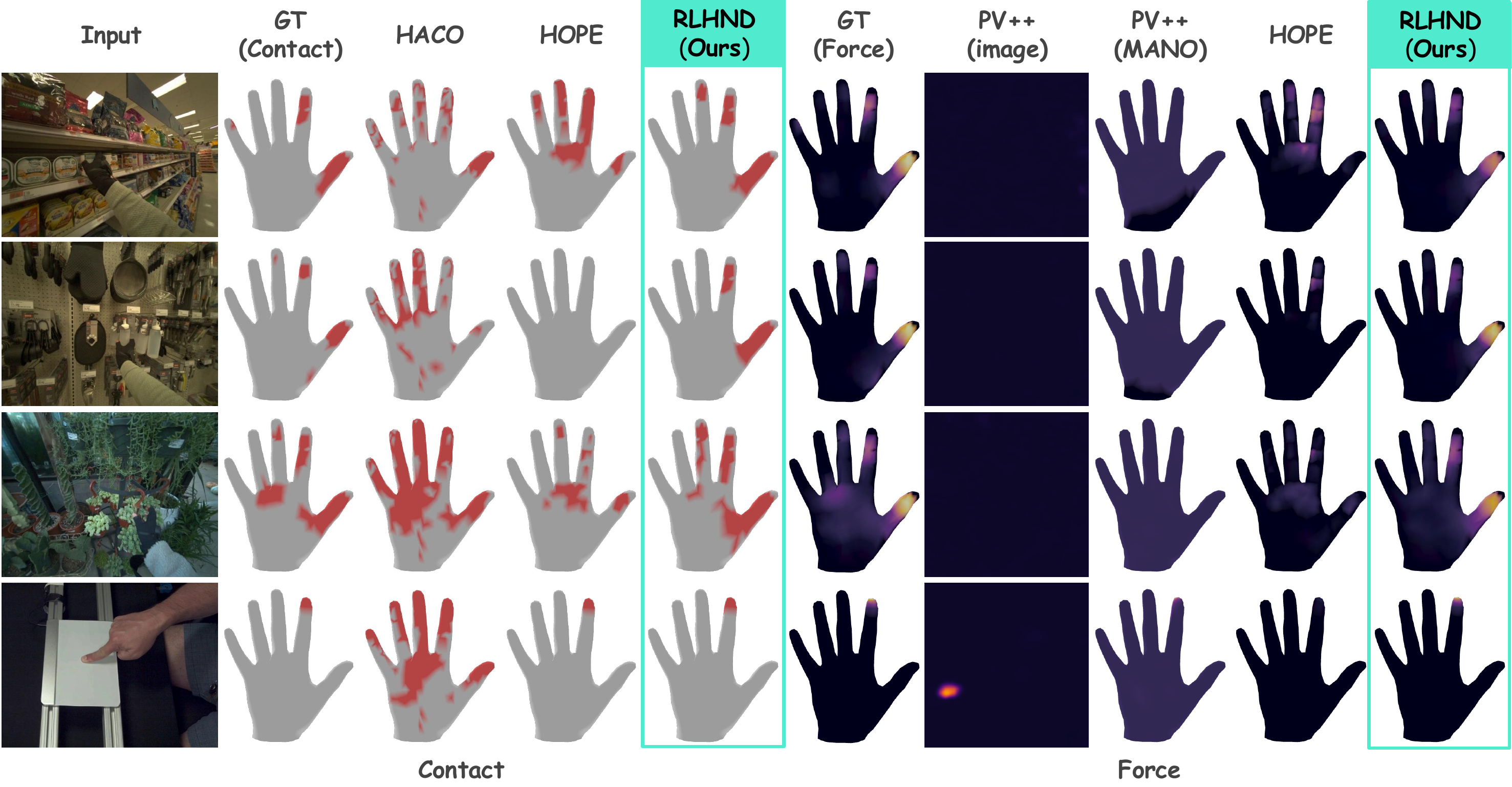}
\caption{
\textbf{Qualitative comparison of contact and force estimation.}
We visualize the \emph{contact} (left) and \emph{force} (right) estimation from OpenTouch (first to third row) and PVDB (fourth row).
}
\label{fig:qualitative_tactile}
\end{figure}

\paragraph{Setup.}
Stage-1 is trained on a weighted mixture of egocentric and exocentric hand-video datasets, and stage-2 on force-labeled glove and pressure-pad recordings together with contact-only labels derived from hand--object meshes, which supervise only the contact head. The datasets and label generation are described in Appendix~\ref{method:dataset}, the schedules and hyperparameters in Appendix~\ref{app:exp_details}, and the metrics in Appendix~\ref{app:metrics}.
Motion reconstruction is evaluated on the held-out HOT3D and ARCTIC egocentric splits and, zero-shot for our method and every baseline, on EgoDex; tactile estimation on the OpenTouch and PressureVisionDB (PVDB) test splits of HOPE~\citep{jeon2026hope} and, for contact only, on HOT3D, DexYCB, and ARCTIC with mesh-derived contact labels.
All baselines use their official weights and inference code.

\subsection{Hand Pose Estimation}
\begin{table}[t]
\centering
\setlength{\tabcolsep}{5pt}
\resizebox{\columnwidth}{!}{%
\begin{tabular}{llccccccc}
\toprule
Test set & Method & F$_{\mathrm{Acc}}$ $\uparrow$ & MPJPE-p $\downarrow$ & PA-p $\downarrow$ & MPJPE$^{+\mathrm{OOS}}$ $\downarrow$ & EPE2D-p $\downarrow$ & Jitter $\downarrow$ & $\sigma_{\mathrm{shape}}$ $\downarrow$ \\
\midrule
\multirow{8}{*}{HOT3D}
& HaMeR~\citep{pavlakos2024reconstructing} & 0.743 & 65.53 & 11.42 & 66.85 & 130.33 & 25.40 & \second{1.17} \\
& WiLoR~\citep{potamias2024wilor} & 0.743 & 44.89 & 9.71 & 46.23 & 134.89 & 22.17 & 1.90 \\
& HaWoR~\citep{zhang2025hawor} & 0.743 & 41.53 & 10.15 & 42.99 & 126.10 & 26.13 & 7.37 \\
& HaPTIC~\citep{ye2025haptic} & 0.743 & 62.42 & 11.26 & 63.70 & 130.40 & 38.27 & 3.20 \\
& HandFlow~\citep{xu2026handflow} & 0.729 & 39.90 & 9.50 & 24.87 & 127.85 & 16.99 & 3.68 \\
& ACE-Ego-Hand~\citep{liu2026aceegohand} & \second{0.928} & 24.41 & 8.60 & 20.30 & 38.09 & 6.48 & 1.22 \\
\cmidrule(lr){2-9}
& \textbf{RLHND (Ours)} & \best{0.996} & \second{13.01} & \second{6.72} & \second{13.08} & \best{6.84} & \best{4.38} & 1.30 \\
& \textbf{RLHND (Ours)} ($\boldsymbol\beta$-cache) & \best{0.996} & \best{13.00} & \best{6.68} & \best{13.07} & \second{6.89} & \second{4.39} & \best{0.00} \\
\midrule
\multirow{8}{*}{ARCTIC ego}
& HaMeR~\citep{pavlakos2024reconstructing} & 0.887 & 31.76 & 9.72 & 52.99 & 86.30 & 21.73 & \second{0.52} \\
& WiLoR~\citep{potamias2024wilor} & 0.887 & 24.05 & 7.49 & 46.33 & 87.66 & 25.92 & 6.15 \\
& HaWoR~\citep{zhang2025hawor} & 0.886 & 31.72 & 9.55 & 52.88 & 82.72 & 29.34 & 5.44 \\
& HaPTIC~\citep{ye2025haptic} & 0.886 & 24.72 & 7.92 & 46.88 & 84.05 & 51.09 & 6.19 \\
& HandFlow~\citep{xu2026handflow} & 0.918 & 40.03 & 9.91 & 47.47 & 74.09 & 13.47 & 5.12 \\
& ACE-Ego-Hand~\citep{liu2026aceegohand} & \best{0.994} & 14.86 & 7.50 & 16.86 & \best{11.76} & \best{3.52} & 1.91 \\
\cmidrule(lr){2-9}
& \textbf{RLHND (Ours)} & \second{0.993} & \best{13.36} & \best{6.42} & \best{15.40} & \second{12.19} & 4.19 & 0.96 \\
& \textbf{RLHND (Ours)} ($\boldsymbol\beta$-cache) & \second{0.993} & \second{13.44} & \second{6.44} & \second{15.46} & 12.28 & \second{4.17} & \best{0.00} \\
\midrule
\multirow{8}{*}{EgoDex}
& HaMeR~\citep{pavlakos2024reconstructing} & 0.970 & 45.97 & 11.79 & 45.90 & 66.31 & 15.48 & 0.43 \\
& WiLoR~\citep{potamias2024wilor} & 0.970 & 42.57 & 10.87 & 42.50 & 67.68 & 13.05 & \second{0.22} \\
& HaWoR~\citep{zhang2025hawor} & 0.970 & 27.76 & 11.22 & 27.73 & 54.20 & 16.94 & 4.58 \\
& HaPTIC~\citep{ye2025haptic} & 0.970 & 40.92 & 11.29 & 40.85 & 65.38 & 25.81 & 2.33 \\
& HandFlow~\citep{xu2026handflow} & 0.978 & 24.60 & 11.45 & 22.34 & 41.44 & 10.89 & 3.57 \\
& ACE-Ego-Hand~\citep{liu2026aceegohand} & \second{0.999} & 20.95 & 10.99 & 20.85 & \second{22.26} & 3.01 & 0.59 \\
\cmidrule(lr){2-9}
& \textbf{RLHND (Ours)} & \best{1.000} & \best{19.90} & \best{10.12} & \best{19.89} & \best{20.72} & \best{2.93} & 0.71 \\
& \textbf{RLHND (Ours)} ($\boldsymbol\beta$-cache) & \best{1.000} & \second{19.93} & \second{10.14} & \second{19.91} & \best{20.72} & \second{2.94} & \best{0.00} \\
\bottomrule
\end{tabular}%
}
\caption{
\textbf{Quantitative comparison of motion reconstruction.}
RLHND outperforms baselines on nearly every metric on every benchmark consistently.
We denote \bestcap{best} and \secondcap{second best} values.
}
\label{tab:pose_comparison}
\end{table}

In Figure~\ref{fig:qualitative_pose}, crop-based models exhibit noisy hand trajectories and incorrect mesh overlays, while ACE-Ego-Hand performs better but still shows erroneous hand detection (the left hand in the first row), unrealistic motion (twisted fingers in the second row), and inaccurate hand scale (the third-row overlay). In contrast, ours produces robust mesh overlays and smooth hand trajectories.

Table~\ref{tab:pose_comparison} reports quantitative comparisons with per-window $\boldsymbol\beta$ estimation and $\boldsymbol\beta$-caching, which estimates $\boldsymbol\beta$ from the first output window and reuses it throughout the video.
Ours consistently outperforms across all test sets, including zero-shot EgoDex, with higher F$_{\mathrm{Acc}}$ than the external hand detectors used by HaMeR, WiLoR, HaWoR, and HaPTIC, as well as strong 3D and 2D projection metrics.
Although $\boldsymbol\beta$-caching removes $\sigma_{\mathrm{shape}}$, it does not improve jitter, as analyzed in Appendix~\ref{app:jitter_beta}.

\subsection{Tactile Estimation}
\begin{table}[t]
\centering
\setlength{\tabcolsep}{3pt}
\resizebox{\columnwidth}{!}{%
\begin{tabular}{lcccccccccccc}
\toprule
& \multicolumn{8}{c}{Contact (vertex-level)} & \multicolumn{4}{c}{Force (kPa)} \\
\cmidrule(lr){2-9}\cmidrule(lr){10-13}
& \multicolumn{2}{c}{OpenTouch} & \multicolumn{2}{c}{DexYCB} & \multicolumn{2}{c}{HOT3D} & \multicolumn{2}{c}{ARCTIC} & \multicolumn{2}{c}{OpenTouch} & \multicolumn{2}{c}{PVDB} \\
\cmidrule(lr){2-3} \cmidrule(lr){4-5} \cmidrule(lr){6-7} \cmidrule(lr){8-9} \cmidrule(lr){10-11} \cmidrule(lr){12-13}
Method & F1 $\uparrow$ & AUROC $\uparrow$ & F1 $\uparrow$ & AUROC $\uparrow$ & F1 $\uparrow$ & AUROC $\uparrow$ & F1 $\uparrow$ & AUROC $\uparrow$ & MAE $\downarrow$ & RMSE $\downarrow$ & MAE $\downarrow$ & RMSE $\downarrow$ \\
\midrule
PressureVision~\citep{grady2022pressurevision} & -- & -- & -- & -- & -- & -- & -- & -- & 1.930 & 6.190 & 0.471 & 3.726 \\
PressureVision++~\citep{grady2024pressurevision} & -- & -- & -- & -- & -- & -- & -- & -- & 1.920 & 6.200 & \second{0.248} & 3.025 \\
HACO~\citep{jung2026learning} & 0.373 & 0.541 & \second{0.543} & \second{0.883} & \second{0.244} & 0.749 & 0.580 & 0.907 & -- & -- & -- & -- \\
HOPE~\citep{jeon2026hope} & \second{0.663} & \second{0.894} & 0.506 & 0.868 & 0.197 & \second{0.762} & \second{0.591} & \second{0.914} & \second{1.781} & \second{5.388} & \best{0.236} & \second{2.091} \\
\midrule
\textbf{RLHND (Ours)} & \best{0.696} & \best{0.980} & \best{0.572} & \best{0.915} & \best{0.589} & \best{0.959} & \best{0.602} & \best{0.935} & \best{0.489} & \best{2.508} & 0.274 & \best{2.066} \\
\bottomrule
\end{tabular}%
}
\caption{
\textbf{Quantitative comparison of contact and force estimation.}
RLHND outperforms baselines on nearly every benchmark consistently.
We denote \bestcap{best} and \secondcap{second best} values.
}
\label{tab:tactile_merged}
\end{table}

Figure~\ref{fig:qualitative_tactile} compares predicted contact and pressure maps on OpenTouch and PressureVisionDB.
RLHND produces localized contact regions and accurate pressure patterns on the hand surface, whereas HACO over-predicts contact and HOPE produces fragmented contact regions with underestimated pressure.
PressureVision++ predicts pressure only in its native planar-sensor representation.

Table~\ref{tab:tactile_merged} quantifies contact (F1 and AUROC) on OpenTouch, HOT3D, DexYCB, and ARCTIC and force on OpenTouch and PVDB; the PressureVision family~\citep{grady2022pressurevision,grady2024pressurevision} predicts pressure on the image plane rather than the hand surface, so its contact metrics are not reported.
RLHND achieves the best contact scores across all four datasets in both metrics, with competitive force estimation.

\subsection{Effectiveness on Robot Learning}
\label{sec:retargeting}
\begin{table}[t]
\centering
\setlength{\tabcolsep}{3pt}
\resizebox{\columnwidth}{!}{%
\begin{tabular}{lcccccccccc}
\toprule
& \multicolumn{2}{c}{Sharpa Wave (22 DoF)} & \multicolumn{2}{c}{WUJI v2 (20 DoF)} & \multicolumn{2}{c}{Shadow (24 DoF)} & \multicolumn{2}{c}{Inspire RH56 (12 DoF)} & \multicolumn{2}{c}{ALLEX (15 DoF)} \\
\cmidrule(lr){2-3}\cmidrule(lr){4-5}\cmidrule(lr){6-7}\cmidrule(lr){8-9}\cmidrule(lr){10-11}
Method & Q-err $\downarrow$ & Jerk $\downarrow$ & Q-err $\downarrow$ & Jerk $\downarrow$ & Q-err $\downarrow$ & Jerk $\downarrow$ & Q-err $\downarrow$ & Jerk $\downarrow$ & Q-err $\downarrow$ & Jerk $\downarrow$ \\
\midrule
\textcolor{gray}{Oracle (GT joints)} & \textcolor{gray}{0.0} & \textcolor{gray}{0.34} & \textcolor{gray}{0.0} & \textcolor{gray}{0.33} & \textcolor{gray}{0.0} & \textcolor{gray}{0.35} & \textcolor{gray}{0.0} & \textcolor{gray}{0.18} & \textcolor{gray}{0.0} & \textcolor{gray}{0.36} \\
HaMeR~\citep{pavlakos2024reconstructing} & 16.6 & 1.45 & 19.4 & 1.31 & 12.6 & 1.31 & 8.1 & 0.90 & 14.1 & 1.19 \\
WiLoR~\citep{potamias2024wilor} & 14.2 & 1.48 & 17.0 & 1.34 & 10.2 & 1.20 & \second{6.3} & 0.90 & 11.2 & 1.18 \\
HaWoR~\citep{zhang2025hawor} & 15.0 & 0.61 & 18.0 & 0.47 & 12.2 & 0.50 & 8.9 & 0.39 & 14.4 & 0.50 \\
HaPTIC~\citep{ye2025haptic} & 14.9 & 1.39 & 17.6 & 1.27 & 11.9 & 1.10 & 7.0 & 0.87 & 12.7 & 1.21 \\
HandFlow~\citep{xu2026handflow} & 18.7 & 0.58 & 21.8 & 0.55 & 14.2 & 0.52 & 9.9 & 0.49 & 17.8 & 0.53 \\
ACE-Ego-Hand~\citep{liu2026aceegohand} & 14.0 & 0.20 & 17.4 & 0.21 & 10.8 & 0.21 & 7.4 & \second{0.11} & 12.0 & \second{0.20} \\
\cmidrule(lr){1-11}
\textbf{RLHND (Ours)} & \best{12.7} & \second{0.18} & \best{16.1} & \best{0.16} & \second{9.6} & \best{0.19} & \best{6.0} & \best{0.07} & \best{10.2} & \best{0.17} \\
\textbf{RLHND (Ours)} ($\boldsymbol\beta$-cache) & \second{12.8} & \best{0.17} & \second{16.3} & \second{0.17} & \best{9.5} & \second{0.20} & \best{6.0} & \best{0.07} & \second{10.3} & \best{0.17} \\
\bottomrule
\end{tabular}%
}
\caption{Retargeting quality on five dexterous right hands (DexPilot, wrist-relative targets), averaged over the HOT3D, ARCTIC ego and EgoDex test sets; per-dataset results are in Appendix~\ref{app:retargeting_full}. Q-err is the mean absolute deviation ($^\circ$) of the joint command from the one the same solver produces on the ground-truth joints, and Jerk the median frame-to-frame command noise ($^\circ$/frame$^2$); see Appendix~\ref{app:metrics}. The grey Oracle row retargets the ground-truth joints.
}\label{tab:retargeting}
\end{table}

\begin{table}[t]
\centering
\small
\setlength{\tabcolsep}{4pt}
\resizebox{\columnwidth}{!}{%
\begin{tabular}{lccccccccccc}
\toprule
\multirow{2}{*}{Method} & \multirow{2}{*}{\begin{tabular}[c]{@{}c@{}}Human : Robot\\episodes\end{tabular}} & \multicolumn{5}{c}{Pick and Place} & \multicolumn{3}{c}{Bimanual} & \multirow{2}{*}{All $\uparrow$} \\
\cmidrule(lr){3-7}\cmidrule(lr){8-10}
& & Bottle & Box & Ball & Bird & Avg.\ $\uparrow$ & Plastic bag & Assemble tissue & Avg.\ $\uparrow$ & \\
\midrule
DPP~\citep{kim2026dexpointpolicy} & 100 : 100 & \best{93.8} & \best{100.0} & 81.3 & 81.3 & 89.1 & 53.1 & 75.0 & 64.1 & 80.8 \\
\textbf{DPP + RLHND (Ours)} & 100 : 100 & \best{93.8} & \best{100.0} & \best{87.5} & \best{93.8} & \best{93.8} & \best{59.4} & \best{90.6} & \best{75.0} & \best{87.5} \\
\bottomrule
\end{tabular}%
}
\caption{
\textbf{Real-robot success rate of Dexterous Point Policy.} DPP is trained on human demonstrations labeled by its original tracker, and DPP + RLHND on the same demonstrations labeled by RLHND.
}
\vspace{-0.5cm}
\label{tab:robot_learning}
\end{table}

\paragraph{Real-robot experiments.}
Dexterous Point Policy~\citep{kim2026dexpointpolicy}, \ie DPP, trains robot policies from human demonstrations using keypoint locations and fingertip contact labels for force application, with the latter \emph{manually} annotated for each demonstration.
RLHND provides both \emph{automatically} from a single model, with fingertip keypoints from its pose stream and contact labels from its contact head. Table~\ref{tab:robot_learning} compares rollout results with DPP + RLHND; since the original tracker predicts no contact, both rows use RLHND's contact labels and differ only in the hand keypoints. DPP + RLHND improves performance across the evaluated tasks, with larger gains on the complex bimanual tasks.

\paragraph{Retargeting to robot hands.}
We evaluate how well tracker outputs serve as robot action spaces in Table~\ref{tab:retargeting}.
We retarget per-frame predictions to five dexterous right hands using the same DexPilot~\citep{handa2020dexpilot} solver with wrist-relative targets, and report Q-err and joint jerk averaged over HOT3D, ARCTIC ego, and EgoDex.
RLHND, with or without the $\boldsymbol\beta$-cache, achieves the lowest mean Q-err across all five embodiments and produces among the smoothest commands.

\subsection{Ablation Studies}
\label{sec:ablation}
\begin{table}[t]
\setlength{\tabcolsep}{4pt}
\noindent\resizebox{\columnwidth}{!}{%
\begin{tabular}{lcccccccc}
\toprule
& \multicolumn{4}{c}{ARCTIC ego} & \multicolumn{4}{c}{EgoDex} \\
\cmidrule(lr){2-5}\cmidrule(lr){6-9}
Pose variant & MPJPE-p $\downarrow$ & PA-p $\downarrow$ & EPE2D-p $\downarrow$ & Q-err $\downarrow$ & MPJPE-p $\downarrow$ & PA-p $\downarrow$ & EPE2D-p $\downarrow$ & Q-err $\downarrow$ \\
\midrule
A0: w/o Cosmos 3 & 16.62 & 7.55 & 13.40 & 12.48 & 21.57 & 10.63 & 22.45 & 15.64 \\
A1: w/o constrained pose & 14.26 & 6.52 & 12.33 & 10.85 & \second{19.91} & 10.38 & \second{20.76} & 15.31 \\
A2: w/o $\boldsymbol\beta$-conditioning & 14.87 & 6.72 & 12.27 & 10.48 & 20.06 & \second{10.33} & 20.95 & \second{15.29} \\
\midrule
A3: Full model & \second{13.36} & \best{6.42} & \best{12.19} & \second{10.12} & \best{19.90} & \best{10.12} & \best{20.72} & \best{15.22} \\
A4: Full model + GT $\boldsymbol\beta$ & \best{13.15} & \second{6.45} & \second{12.21} & \best{9.69} & -- & -- & -- & -- \\
\bottomrule
\end{tabular}%
}\par\vspace{3pt}
\noindent\resizebox{\columnwidth}{!}{%
\begin{tabular}{lccccccc c}
\toprule
& \multicolumn{2}{c}{DexYCB} & \multicolumn{2}{c}{HOT3D} & \multicolumn{2}{c}{ARCTIC} & \multicolumn{2}{c}{OpenTouch force (kPa)} \\
\cmidrule(lr){2-3}\cmidrule(lr){4-5}\cmidrule(lr){6-7}\cmidrule(lr){8-9}
Tactile variant & F1 $\uparrow$ & AUROC $\uparrow$ & F1 $\uparrow$ & AUROC $\uparrow$ & F1 $\uparrow$ & AUROC $\uparrow$ & MAE $\downarrow$ & RMSE $\downarrow$ \\
\midrule
B0: w/o Cosmos 3 & \second{0.549} & \second{0.900} & 0.548 & 0.940 & \second{0.557} & \second{0.919} & \second{0.490} & 2.686 \\
B1: w/o LBS spread & 0.532 & 0.892 & \second{0.569} & \second{0.948} & \second{0.557} & 0.913 & 0.537 & 2.623 \\
B2: w/o contact-only data & 0.047 & 0.683 & 0.145 & 0.616 & 0.117 & 0.755 & 0.500 & \second{2.524} \\
\midrule
B3: Full tactile expert & \best{0.572} & \best{0.915} & \best{0.589} & \best{0.959} & \best{0.602} & \best{0.935} & \best{0.489} & \best{2.508} \\
\bottomrule
\end{tabular}%
}\par
\caption{
\textbf{Leave-one-out ablations of RLHND.}
\emph{Top (pose):} variants A0-A4, evaluated on ARCTIC ego and EgoDex.
\emph{Bottom (tactile):} variants B0--B3, evaluated on contact benchmarks and OpenTouch.
}
\vspace{-0.5cm}
\label{tab:ablation}
\end{table}

We ablate the three components that distinguish RLHND from its base tracker by removing each from the full model and retraining under the same 20k-step schedule and data mixture (Table~\ref{tab:ablation}). Variant (A0) replaces the Cosmos~3 Nano encoder with the Wan2.2 backbone following ACE-Ego-Hand, (A1) replaces the anatomically constrained parameterization of Section~\ref{method:decoder} with 16 free 6D rotations supervised on the raw 45-DoF MANO labels, and (A2) disables external-$\boldsymbol\beta$ injection and teacher forcing, so shape is predicted by the shape head alone, without the one-time hand calibration of Section~\ref{method:decoder}, as in the full model (A3).

The encoder is the dominant factor, with A0 degrading performance across all metrics and test sets, confirming that a strong video backbone provides substantial gains.
Removing either the anatomically constrained pose representation (A1) or shape caching (A2) also degrades MPJPE in ARCTIC, while their combination yields a larger improvement in the full model. The full model with ground-truth $\boldsymbol\beta$ (A4) gains nothing in joint-level metrics but improves Q-err, suggesting that fixed hand shape may stabilize retargeting.

Each tactile variant retrains the stage-2 expert on the frozen full pose model: B0 uses the Wan2.2 stage-1 model of A0, B1 replaces the fixed MANO skinning weights $\mathbf{W}_{\mathbf{lbs}}$ of Eq.~(\ref{eq:lbs_spread}) with a learnable $778{\times}16$ spread matrix, and B2 trains on OpenTouch and PVDB alone. B0 and B1 both degrade performance, while removing the contact-only data (B2) causes a far larger contact drop with no notable change in force.

\section{Conclusion}
\label{sec:conclusion}
We presented RLHND, a physically grounded hand tracker built on a pre-trained video foundation model for robot learning.
It achieves state-of-the-art motion reconstruction and tactile estimation, produces smooth retargeted commands across five dexterous hands, and lets Dexterous Point Policy outperform its original tracker without manual annotation.

\section*{Reproducibility Statement}
The architecture and training procedure are specified in Section~\ref{method:encoder}--\ref{method:training}, with implementation details, label construction, losses, evaluation metrics, training schedules, and baseline protocols provided in Appendix~\ref{app:impl}.
All training and evaluation data come from publicly available datasets, and the retargeting solver and its settings are described in Section~\ref{sec:retargeting}.
We will release the implementation of RLHND, along with the pre-trained weights.

\section*{Ethics Statement}
RLHND uses only publicly released human hand-object datasets under their respective licenses and consent procedures.
We collect no new human data and release no raw video.
Since the method can recover fine-grained hand activity from ordinary footage, including egocentric recordings containing bystanders, we intend it for robot learning from consenting demonstrators and discourage surveillance use.
Contact and force are inferred rather than measured, so we recommend joint-torque or current limits and human supervision when deploying retargeted commands.
Finally, the limited diversity of public datasets may lead to performance variation across subjects, hand sizes, and skin tones that is not captured by our evaluation sets. The authors declare no conflicts of interest.

\section*{AI Use Statement}
We used generative AI tools to assist with figure and table editing, and routine coding.
AI-assisted code was reviewed and tested by the authors, and all reported results were independently produced and verified by the authors.
All AI-assisted text and figures were reviewed against the underlying experiments and data.
The authors take full responsibility for the content of this paper.

\bibliography{realhand}

\begin{thebibliography}{61}
\providecommand{\natexlab}[1]{#1}
\providecommand{\url}[1]{\texttt{#1}}
\expandafter\ifx\csname urlstyle\endcsname\relax
  \providecommand{\doi}[1]{doi: #1}\else
  \providecommand{\doi}{doi: \begingroup \urlstyle{rm}\Url}\fi

\bibitem[Bai et~al.(2025)Bai, Chen, Liu, Wang, Ge, Song, Dang, Wang, Wang, Tang, et~al.]{bai2025qwen25vl}
Shuai Bai, Keqin Chen, Xuejing Liu, Jialin Wang, Wenbin Ge, Sibo Song, Kai Dang, Peng Wang, Shijie Wang, Jun Tang, et~al.
\newblock {Qwen2.5-VL} technical report.
\newblock \emph{arXiv preprint arXiv:2502.13923}, 2025.

\bibitem[Banerjee et~al.(2024)Banerjee, Shkodrani, Moulon, Hampali, Zhang, Fountain, Miller, Basol, Newcombe, Wang, et~al.]{banerjee2024introducing}
Prithviraj Banerjee, Sindi Shkodrani, Pierre Moulon, Shreyas Hampali, Fan Zhang, Jade Fountain, Edward Miller, Selen Basol, Richard Newcombe, Robert Wang, et~al.
\newblock Introducing {HOT3D}: An egocentric dataset for {3D} hand and object tracking.
\newblock \emph{arXiv preprint arXiv:2406.09598}, 2024.

\bibitem[Bjorck et~al.(2025)Bjorck, Casta{\~n}eda, Cherniadev, Da, Ding, Fan, Fang, Fox, Hu, Huang, et~al.]{bjorck2025gr00t}
Johan Bjorck, Fernando Casta{\~n}eda, Nikita Cherniadev, Xingye Da, Runyu Ding, Linxi Fan, Yu~Fang, Dieter Fox, Fengyuan Hu, Spencer Huang, et~al.
\newblock {GR00T N1}: An open foundation model for generalist humanoid robots.
\newblock \emph{arXiv preprint arXiv:2503.14734}, 2025.

\bibitem[Black et~al.(2024)Black, Brown, Driess, Esmail, Equi, Finn, Fusai, Groom, Hausman, Ichter, et~al.]{black2024pi0}
Kevin Black, Noah Brown, Danny Driess, Adnan Esmail, Michael Equi, Chelsea Finn, Niccolo Fusai, Lachy Groom, Karol Hausman, Brian Ichter, et~al.
\newblock $\pi_0$: A vision-language-action flow model for general robot control.
\newblock \emph{arXiv preprint arXiv:2410.24164}, 2024.

\bibitem[Brohan et~al.(2023)Brohan, Brown, Carbajal, Chebotar, Chen, Choromanski, Ding, Driess, Dubey, Finn, et~al.]{brohan2023rt2}
Anthony Brohan, Noah Brown, Justice Carbajal, Yevgen Chebotar, Xi~Chen, Krzysztof Choromanski, Tianli Ding, Danny Driess, Avinava Dubey, Chelsea Finn, et~al.
\newblock {RT-2}: Vision-language-action models transfer web knowledge to robotic control.
\newblock \emph{arXiv preprint arXiv:2307.15818}, 2023.

\bibitem[Cai et~al.(2025)Cai, Qiu, Chen, Wei, Liu, Huang, Cheng, and Wang]{cai2025innon}
Xiongyi Cai, Ri-Zhao Qiu, Geng Chen, Lai Wei, Isabella Liu, Tianshu Huang, Xuxin Cheng, and Xiaolong Wang.
\newblock {In-N-On}: Scaling egocentric manipulation with in-the-wild and on-task data.
\newblock \emph{arXiv preprint arXiv:2511.15704}, 2025.

\bibitem[Carion et~al.(2025)Carion, Gustafson, Hu, Debnath, Hu, Suris, Ryali, Alwala, Khedr, Huang, et~al.]{carion2025sam3}
Nicolas Carion, Laura Gustafson, Yuan-Ting Hu, Shoubhik Debnath, Ronghang Hu, Didac Suris, Chaitanya Ryali, Kalyan~Vasudev Alwala, Haitham Khedr, Andrew Huang, et~al.
\newblock {SAM} 3: Segment anything with concepts.
\newblock \emph{arXiv preprint arXiv:2511.16719}, 2025.

\bibitem[Chao et~al.(2021)Chao, Yang, Xiang, Molchanov, Handa, Tremblay, Narang, Van~Wyk, Iqbal, Birchfield, et~al.]{chao2021dexycb}
Yu-Wei Chao, Wei Yang, Yu~Xiang, Pavlo Molchanov, Ankur Handa, Jonathan Tremblay, Yashraj~S Narang, Karl Van~Wyk, Umar Iqbal, Stan Birchfield, et~al.
\newblock {DexYCB}: A benchmark for capturing hand grasping of objects.
\newblock In \emph{Proceedings of the IEEE/CVF conference on computer vision and pattern recognition}, pp.\  9044--9053, 2021.

\bibitem[Chen et~al.(2026)Chen, Wan, Chen, Guo, Xu, Qi, Zhang, Wu, Xu, Li, Wu, Li, Yang, Luo, Sui, and Mu]{chen2026univtac}
Baijun Chen, Weijie Wan, Tianxing Chen, Xianda Guo, Congsheng Xu, Yuanyang Qi, Haojie Zhang, Longyan Wu, Tianling Xu, Zixuan Li, Yizhe Wu, Rui Li, Xiaokang Yang, Ping Luo, Wei Sui, and Yao Mu.
\newblock {UniVTAC}: A unified simulation platform for visuo-tactile manipulation data generation, learning, and benchmarking.
\newblock \emph{arXiv preprint arXiv:2602.10093}, 2026.

\bibitem[Fan et~al.(2023)Fan, Taheri, Tzionas, Kocabas, Kaufmann, Black, and Hilliges]{fan2023arctic}
Zicong Fan, Omid Taheri, Dimitrios Tzionas, Muhammed Kocabas, Manuel Kaufmann, Michael~J Black, and Otmar Hilliges.
\newblock {ARCTIC}: A dataset for dexterous bimanual hand-object manipulation.
\newblock In \emph{Proceedings of the IEEE/CVF Conference on Computer Vision and Pattern Recognition}, pp.\  12943--12954, 2023.

\bibitem[{Gemini Team}(2023)]{team2023gemini}
{Gemini Team}.
\newblock Gemini: A family of highly capable multimodal models.
\newblock \emph{arXiv preprint arXiv:2312.11805}, 2023.

\bibitem[Grady et~al.(2022)Grady, Tang, Brahmbhatt, Twigg, Wan, Hays, and Kemp]{grady2022pressurevision}
Patrick Grady, Chengcheng Tang, Samarth Brahmbhatt, Christopher~D Twigg, Chengde Wan, James Hays, and Charles~C Kemp.
\newblock {PressureVision}: Estimating hand pressure from a single {RGB} image.
\newblock In \emph{European Conference on Computer Vision (ECCV)}, 2022.

\bibitem[Grady et~al.(2024)Grady, Zhao, Tang, Kumar, Twigg, Ye, Vouga, and Kemp]{grady2024pressurevision}
Patrick Grady, Jeremy~A Zhao, Chengcheng Tang, Aditya Kumar, Christopher~D Twigg, Yafei Ye, Etienne Vouga, and Charles~C Kemp.
\newblock {PressureVision++}: Estimating fingertip pressure from diverse {RGB} images.
\newblock In \emph{Proceedings of the IEEE/CVF Winter Conference on Applications of Computer Vision}, pp.\  8698--8708, 2024.

\bibitem[Grauman et~al.(2022)Grauman, Westbury, Byrne, Chavis, Furnari, Girdhar, Hamburger, Jiang, Liu, Liu, et~al.]{grauman2022ego4d}
Kristen Grauman, Andrew Westbury, Eugene Byrne, Zachary Chavis, Antonino Furnari, Rohit Girdhar, Jackson Hamburger, Hao Jiang, Miao Liu, Xingyu Liu, et~al.
\newblock {Ego4D}: Around the world in 3,000 hours of egocentric video.
\newblock In \emph{Proceedings of the IEEE/CVF Conference on Computer Vision and Pattern Recognition}, 2022.

\bibitem[Haldar \& Pinto(2025)Haldar and Pinto]{haldar2025pointpolicy}
Siddhant Haldar and Lerrel Pinto.
\newblock Point policy: Unifying observations and actions with key points for robot manipulation.
\newblock In \emph{Conference on Robot Learning}, 2025.

\bibitem[Hampali et~al.(2020)Hampali, Rad, Oberweger, and Lepetit]{hampali2020honnotate}
Shreyas Hampali, Mahdi Rad, Markus Oberweger, and Vincent Lepetit.
\newblock Honnotate: A method for {3D} annotation of hand and object poses.
\newblock In \emph{Proceedings of the IEEE/CVF Conference on Computer Vision and Pattern Recognition}, pp.\  3196--3206, 2020.

\bibitem[Handa et~al.(2020)Handa, Van~Wyk, Yang, Liang, Chao, Wan, Birchfield, Ratliff, and Fox]{handa2020dexpilot}
Ankur Handa, Karl Van~Wyk, Wei Yang, Jacky Liang, Yu-Wei Chao, Qian Wan, Stan Birchfield, Nathan Ratliff, and Dieter Fox.
\newblock {DexPilot}: Vision-based teleoperation of dexterous robotic hand-arm system.
\newblock In \emph{IEEE International Conference on Robotics and Automation}, 2020.

\bibitem[Hoque et~al.(2026)Hoque, Huang, Yoon, Sivapurapu, and Zhang]{hoque2026egodex}
Ryan Hoque, Peide Huang, David~J. Yoon, Mouli Sivapurapu, and Jian Zhang.
\newblock {EgoDex}: Learning dexterous manipulation from large-scale egocentric video.
\newblock In \emph{International Conference on Learning Representations}, 2026.

\bibitem[Hu et~al.(2022)Hu, Shen, Wallis, Allen-Zhu, Li, Wang, Wang, and Chen]{hu2022lora}
Edward~J Hu, Yelong Shen, Phillip Wallis, Zeyuan Allen-Zhu, Yuanzhi Li, Shean Wang, Lu~Wang, and Weizhu Chen.
\newblock Lo{RA}: Low-rank adaptation of large language models.
\newblock In \emph{International Conference on Learning Representations}, 2022.

\bibitem[Jeon et~al.(2026)Jeon, Kim, and Joo]{jeon2026hope}
Subin Jeon, Byungjun Kim, and Hanbyul Joo.
\newblock {HOPE}: Hand-object pressure estimation from monocular videos.
\newblock \emph{arXiv preprint arXiv:2608.06192}, 2026.

\bibitem[Jung \& Lee(2025)Jung and Lee]{jung2026learning}
Daniel Jung and Kyoung~Mu Lee.
\newblock Learning dense hand contact estimation from imbalanced data.
\newblock \emph{Advances in Neural Information Processing Systems}, 38:\penalty0 120351--120384, 2025.

\bibitem[Kareer et~al.(2024)Kareer, Patel, Punamiya, Mathur, Cheng, Wang, Hoffman, and Xu]{kareer2024egomimic}
Simar Kareer, Dhruv Patel, Ryan Punamiya, Pranay Mathur, Shuo Cheng, Chen Wang, Judy Hoffman, and Danfei Xu.
\newblock {EgoMimic}: Scaling imitation learning via egocentric video.
\newblock \emph{arXiv preprint arXiv:2410.24221}, 2024.

\bibitem[Kim et~al.(2026{\natexlab{a}})Kim, Park, Sim, Moon, Lee, and Shin]{kim2026dexpointpolicy}
Beomjun Kim, Seong~Hyeon Park, Seunghoon Sim, Seungjun Moon, Sanghyeok Lee, and Jinwoo Shin.
\newblock Dexterous point policy: Learning point-based dexterous hand policies from human demonstrations.
\newblock \emph{arXiv preprint arXiv:2606.10614}, 2026{\natexlab{a}}.

\bibitem[Kim et~al.(2026{\natexlab{b}})Kim, Jang, Koo, et~al.]{kim2026rldx1}
Dongyoung Kim, Huiwon Jang, Myungkyu Koo, et~al.
\newblock {RLDX-1} technical report.
\newblock \emph{arXiv preprint arXiv:2605.03269}, 2026{\natexlab{b}}.

\bibitem[Kim et~al.(2024)Kim, Pertsch, Karamcheti, Xiao, Balakrishna, Nair, Rafailov, Foster, Lam, Sanketi, et~al.]{kim2024openvla}
Moo~Jin Kim, Karl Pertsch, Siddharth Karamcheti, Ted Xiao, Ashwin Balakrishna, Suraj Nair, Rafael Rafailov, Ethan Foster, Grace Lam, Pannag Sanketi, et~al.
\newblock {OpenVLA}: An open-source vision-language-action model.
\newblock \emph{arXiv preprint arXiv:2406.09246}, 2024.

\bibitem[Kwon et~al.(2021)Kwon, Tekin, St{\"u}hmer, Bogo, and Pollefeys]{kwon2021h2o}
Taein Kwon, Bugra Tekin, Jan St{\"u}hmer, Federica Bogo, and Marc Pollefeys.
\newblock {H2O}: Two hands manipulating objects for first person interaction recognition.
\newblock In \emph{Proceedings of the IEEE/CVF International Conference on Computer Vision}, pp.\  10138--10148, 2021.

\bibitem[Lepert et~al.(2025)Lepert, Fang, and Bohg]{lepert2025phantom}
Marion Lepert, Jiaying Fang, and Jeannette Bohg.
\newblock Phantom: Training robots without robots using only human videos.
\newblock In \emph{Conference on Robot Learning}, 2025.

\bibitem[Li et~al.(2021)Li, Xu, Chen, Bian, Yang, and Lu]{li2021hybrik}
Jiefeng Li, Chao Xu, Zhicun Chen, Siyuan Bian, Lixin Yang, and Cewu Lu.
\newblock {HybrIK}: A hybrid analytical-neural inverse kinematics solution for 3d human pose and shape estimation.
\newblock In \emph{Proceedings of the IEEE/CVF Conference on Computer Vision and Pattern Recognition}, pp.\  3383--3393, 2021.

\bibitem[Li et~al.(2025)Li, Deng, Liang, Luo, Zhou, Yao, Zeng, Feng, Liang, Xu, Zhang, Chen, Chen, Sun, Chen, Yang, and Guo]{li2025vitra}
Qixiu Li, Yu~Deng, Yaobo Liang, Lin Luo, Lei Zhou, Chengtang Yao, Lingqi Zeng, Zhiyuan Feng, Huizhi Liang, Sicheng Xu, Yizhong Zhang, Xi~Chen, Hao Chen, Lily Sun, Dong Chen, Jiaolong Yang, and Baining Guo.
\newblock {VITRA}: Scalable vision-language-action model pretraining for robotic manipulation with real-life human activity videos.
\newblock \emph{arXiv preprint arXiv:2510.21571}, 2025.

\bibitem[Lim et~al.(2026)Lim, Ha, Choi, Kim, Kim, Jeon, and Joo]{lim2026hrdexdb}
Jongbin Lim, Taeyun Ha, Mingi Choi, Jisoo Kim, Byungjun Kim, Subin Jeon, and Hanbyul Joo.
\newblock {HRDexDB}: A paired human-robot dataset for cross-embodiment dexterous grasping.
\newblock \emph{arXiv preprint arXiv:2604.14944}, 2026.

\bibitem[Liu et~al.(2023)Liu, Li, Wu, and Lee]{liu2023visual}
Haotian Liu, Chunyuan Li, Qingyang Wu, and Yong~Jae Lee.
\newblock Visual instruction tuning.
\newblock In \emph{Advances in Neural Information Processing Systems}, 2023.

\bibitem[Liu et~al.(2025)Liu, Adeniji, Zhan, Haldar, Bhirangi, Abbeel, and Pinto]{liu2025egozero}
Vincent Liu, Ademi Adeniji, Haotian Zhan, Siddhant Haldar, Raunaq Bhirangi, Pieter Abbeel, and Lerrel Pinto.
\newblock {EgoZero}: Robot learning from smart glasses.
\newblock \emph{arXiv preprint arXiv:2505.20290}, 2025.

\bibitem[Liu et~al.(2026)Liu, Wang, Li, Zhao, Cai, Jin, Liu, Liu, Huang, Pan, and Li]{liu2026aceegohand}
Yufei Liu, Xixi Wang, Hao Li, Ganlong Zhao, Kaitong Cai, Chengkai Jin, Chunxiao Liu, Jianbo Liu, Siyuan Huang, Xingang Pan, and Hongsheng Li.
\newblock {ACE-Ego-Hand}: Repurposing video diffusion models for occlusion-robust egocentric 3d hand motion recovery, 2026.
\newblock URL \url{https://arxiv.org/abs/2608.20308}.

\bibitem[Luo et~al.(2025)Luo, Feng, Zhang, Zheng, Wang, Yuan, Liu, Xu, Jin, and Lu]{luo2025beingh0}
Hao Luo, Yicheng Feng, Wanpeng Zhang, Sipeng Zheng, Ye~Wang, Haoqi Yuan, Jiazheng Liu, Chaoyi Xu, Qin Jin, and Zongqing Lu.
\newblock {Being-H0}: Vision-language-action pretraining from large-scale human videos.
\newblock \emph{arXiv preprint arXiv:2507.15597}, 2025.

\bibitem[Niu et~al.(2026)Niu, Liu, Wang, Shao, Yin, Pai, Sharma, Saravalle, Zheng, Wang, Punamiya, Xu, Xie, Jiang, Fu, Kallidromitis, Gioia, Zhang, Ge, Feng, Galasso, Zhan, Chan, Bai, Herzig, Lei, Fei-Fei, Goldberg, Malik, Abbeel, Zhu, Xu, Fan, and Darrell]{niu2026trex}
Dantong Niu, Zhuoyang Liu, Zekai Wang, Boning Shao, Zhao-Heng Yin, Anirudh Pai, Yuvan Sharma, Stefano Saravalle, Ruijie Zheng, Jing Wang, Ryan Punamiya, Mengda Xu, Yuqi Xie, Yunfan Jiang, Letian Fu, Konstantinos Kallidromitis, Matteo Gioia, Junyi Zhang, Jiaxin Ge, Haiwen Feng, Fabio Galasso, Wei Zhan, David~M. Chan, Yutong Bai, Roei Herzig, Jiahui Lei, Li~Fei-Fei, Ken Goldberg, Jitendra Malik, Pieter Abbeel, Yuke Zhu, Danfei Xu, Linxi Fan, and Trevor Darrell.
\newblock {T-Rex}: Tactile-reactive dexterous manipulation.
\newblock \emph{arXiv preprint arXiv:2606.17055}, 2026.

\bibitem[{NVIDIA}(2026)]{nvidia2026cosmos3}
{NVIDIA}.
\newblock Cosmos 3: A mixture-of-transformers omni foundation model for physical ai.
\newblock \emph{arXiv preprint arXiv:2606.02800}, 2026.

\bibitem[{Octo Model Team} et~al.(2024){Octo Model Team}, Ghosh, Walke, Pertsch, Black, Mees, Dasari, Hejna, Kreiman, Xu, et~al.]{team2024octo}
{Octo Model Team}, Dibya Ghosh, Homer Walke, Karl Pertsch, Kevin Black, Oier Mees, Sudeep Dasari, Joey Hejna, Tobias Kreiman, Charles Xu, et~al.
\newblock Octo: An open-source generalist robot policy.
\newblock \emph{arXiv preprint arXiv:2405.12213}, 2024.

\bibitem[{OpenAI}(2023)]{openai2023gpt4}
{OpenAI}.
\newblock {GPT-4} technical report.
\newblock \emph{arXiv preprint arXiv:2303.08774}, 2023.

\bibitem[Pavlakos et~al.(2024)Pavlakos, Shan, Radosavovic, Kanazawa, Fouhey, and Malik]{pavlakos2024reconstructing}
Georgios Pavlakos, Dandan Shan, Ilija Radosavovic, Angjoo Kanazawa, David Fouhey, and Jitendra Malik.
\newblock Reconstructing hands in 3d with transformers.
\newblock In \emph{Proceedings of the IEEE/CVF Conference on Computer Vision and Pattern Recognition}, pp.\  9826--9836, 2024.

\bibitem[{Physical Intelligence} et~al.(2025){Physical Intelligence}, Black, Brown, Darpinian, Dhabalia, Driess, Esmail, Equi, Finn, Fusai, et~al.]{intelligence2025pi05}
{Physical Intelligence}, Kevin Black, Noah Brown, James Darpinian, Karan Dhabalia, Danny Driess, Adnan Esmail, Michael Equi, Chelsea Finn, Niccolo Fusai, et~al.
\newblock $\pi_{0.5}$: A vision-language-action model with open-world generalization.
\newblock \emph{arXiv preprint arXiv:2504.16054}, 2025.

\bibitem[Potamias et~al.(2024)Potamias, Zhang, Deng, and Zafeiriou]{potamias2024wilor}
Rolandos~Alexandros Potamias, Jinglei Zhang, Jiankang Deng, and Stefanos Zafeiriou.
\newblock Wilor: End-to-end 3d hand localization and reconstruction in-the-wild, 2024.

\bibitem[Qin et~al.(2023)Qin, Yang, Huang, Van~Wyk, Su, Wang, Chao, and Fox]{qin2023anyteleop}
Yuzhe Qin, Wei Yang, Binghao Huang, Karl Van~Wyk, Hao Su, Xiaolong Wang, Yu-Wei Chao, and Dieter Fox.
\newblock {AnyTeleop}: A general vision-based dexterous robot arm-hand teleoperation system.
\newblock In \emph{Robotics: Science and Systems}, 2023.

\bibitem[Romero et~al.(2017)Romero, Tzionas, and Black]{romero2017embodied}
Javier Romero, Dimitrios Tzionas, and Michael~J Black.
\newblock Embodied hands: Modeling and capturing hands and bodies together.
\newblock \emph{ACM Transactions on Graphics}, 36\penalty0 (6):\penalty0 1--17, 2017.

\bibitem[Shaw et~al.(2022)Shaw, Bahl, and Pathak]{shaw2022videodex}
Kenneth Shaw, Shikhar Bahl, and Deepak Pathak.
\newblock {VideoDex}: Learning dexterity from internet videos.
\newblock In \emph{Conference on Robot Learning}, 2022.

\bibitem[Song et~al.(2025)Song, Li, Fu, Murphy, Zhou, Shiv, Li, Xiong, Owens, Du, et~al.]{song2025opentouch}
Yuxin~Ray Song, Jinzhou Li, Rao Fu, Devin Murphy, Kaichen Zhou, Rishi Shiv, Yaqi Li, Haoyu Xiong, Crystal~Elaine Owens, Yilun Du, et~al.
\newblock {OpenTouch}: Bringing full-hand touch to real-world interaction.
\newblock \emph{arXiv preprint arXiv:2512.16842}, 2025.

\bibitem[{Wan Team}(2025)]{wan2025wan}
{Wan Team}.
\newblock Wan: Open and advanced large-scale video generative models.
\newblock \emph{arXiv preprint arXiv:2503.20314}, 2025.

\bibitem[Wang et~al.(2026)Wang, Jin, Liu, Ouyang, Wei, Zeng, Huang, Shen, and Pan]{wang2026surprising}
Yuxi Wang, Chengkai Jin, Yufei Liu, Wenqi Ouyang, Tianyi Wei, Zhiwei Zeng, Siyuan Huang, Zhiqi Shen, and Xingang Pan.
\newblock The surprising effectiveness of video diffusion models for hand motion reconstruction.
\newblock \emph{arXiv preprint arXiv:2606.30308}, 2026.

\bibitem[Wen et~al.(2025{\natexlab{a}})Wen, Dewan, and Birchfield]{wen2025fastfoundationstereo}
Bowen Wen, Shaurya Dewan, and Stan Birchfield.
\newblock Fast-foundationstereo: Real-time zero-shot stereo matching.
\newblock \emph{arXiv preprint arXiv:2512.11130}, 2025{\natexlab{a}}.

\bibitem[Wen et~al.(2025{\natexlab{b}})Wen, Trepte, Aribido, Kautz, Gallo, and Birchfield]{wen2025foundationstereo}
Bowen Wen, Matthew Trepte, Joseph Aribido, Jan Kautz, Orazio Gallo, and Stan Birchfield.
\newblock Foundationstereo: Zero-shot stereo matching.
\newblock In \emph{Proceedings of the Computer Vision and Pattern Recognition Conference}, 2025{\natexlab{b}}.

\bibitem[Xu et~al.(2026)]{xu2026handflow}
Mingxin Xu et~al.
\newblock Handflow: Fully generative 4d hand recovery with flow matching.
\newblock \url{https://github.com/mxxu00/HandFlow}, 2026.
\newblock V1 open-source release.

\bibitem[Yang et~al.(2021)Yang, Zhan, Li, Xu, Li, and Lu]{yang2021cpf}
Lixin Yang, Xinyu Zhan, Kailin Li, Wenqiang Xu, Jiefeng Li, and Cewu Lu.
\newblock {CPF}: Learning a contact potential field to model the hand-object interaction.
\newblock In \emph{Proceedings of the IEEE/CVF International Conference on Computer Vision}, pp.\  11097--11106, 2021.

\bibitem[Yang et~al.(2025)Yang, Yu, Wu, Yan, Li, Cheng, Zou, Fang, Yin, Liu, et~al.]{yang2025egovla}
Ruihan Yang, Qinxi Yu, Yecheng Wu, Rui Yan, Borui Li, An-Chieh Cheng, Xueyan Zou, Yunhao Fang, Hongxu Yin, Sifei Liu, et~al.
\newblock {EgoVLA}: Learning vision-language-action models from egocentric human videos.
\newblock \emph{arXiv preprint arXiv:2507.12440}, 2025.

\bibitem[Ye et~al.(2025{\natexlab{a}})Ye, Jang, Jeon, Joo, Yang, Peng, Mandlekar, Tan, Chao, Lin, et~al.]{ye2025lapa}
Seonghyeon Ye, Joel Jang, Byeongguk Jeon, Sejune Joo, Jianwei Yang, Baolin Peng, Ajay Mandlekar, Reuben Tan, Yu-Wei Chao, Bill~Yuchen Lin, et~al.
\newblock Latent action pretraining from videos.
\newblock In \emph{International Conference on Learning Representations}, 2025{\natexlab{a}}.

\bibitem[Ye et~al.(2025{\natexlab{b}})Ye, Feng, Taheri, Feng, Tulsiani, and Black]{ye2025haptic}
Yufei Ye, Yao Feng, Omid Taheri, Haiwen Feng, Shubham Tulsiani, and Michael~J. Black.
\newblock Predicting {4D} hand trajectory from monocular videos, 2025{\natexlab{b}}.

\bibitem[Zeng et~al.(2026)Zeng, Shi, Tan, Li, Qin, Lu, Yang, Xue, and Liao]{zeng2026egotactile}
Yuan Zeng, Yujia Shi, Tiao Tan, Xingting Li, Yaqi Qin, Zongqing Lu, Wenming Yang, Jing-Hao Xue, and Qingmin Liao.
\newblock {EgoTactile}: Learning grasp pressure for everyday objects from egocentric video.
\newblock In \emph{International Conference on Machine Learning}, 2026.

\bibitem[Zhan et~al.(2024)Zhan, Yang, Zhao, Mao, Xu, Lin, Li, and Lu]{zhan2024oakink2}
Xinyu Zhan, Lixin Yang, Yifei Zhao, Kangrui Mao, Hanlin Xu, Zenan Lin, Kailin Li, and Cewu Lu.
\newblock {OakInk2}: A dataset of bimanual hands-object manipulation in complex task completion.
\newblock In \emph{Proceedings of the IEEE/CVF Conference on Computer Vision and Pattern Recognition}, pp.\  445--456, 2024.

\bibitem[Zhang et~al.(2026{\natexlab{a}})Zhang, Cai, Xi, Yuan, Luo, Zhang, Zheng, Xu, and Lu]{zhang2026human}
Chi Zhang, Penglin Cai, Ziheng Xi, Haoqi Yuan, Hao Luo, Wanpeng Zhang, Sipeng Zheng, Chaoyi Xu, and Zongqing Lu.
\newblock Human-centric transferable tactile pre-training for dexterous robotic manipulation.
\newblock \emph{arXiv preprint arXiv:2607.01067}, 2026{\natexlab{a}}.

\bibitem[Zhang et~al.(2025)Zhang, Deng, Ma, and Potamias]{zhang2025hawor}
Jinglei Zhang, Jiankang Deng, Chao Ma, and Rolandos~Alexandros Potamias.
\newblock {HaWoR}: World-space hand motion reconstruction from egocentric videos.
\newblock In \emph{Proceedings of the Computer Vision and Pattern Recognition Conference}, pp.\  1805--1815, 2025.

\bibitem[Zhang et~al.(2026{\natexlab{b}})Zhang, Zhang, Shi, Zhu, Zhu, Wang, Wu, and Yuan]{zhang2026unitacvla}
Xidong Zhang, Yichi Zhang, Jiaxin Shi, Fucai Zhu, Siyu Zhu, Michael~Yu Wang, Xiaojun Wu, and Weihao Yuan.
\newblock {UniTacVLA}: Unified tactile understanding and prediction in vision language action models.
\newblock \emph{arXiv preprint arXiv:2606.31723}, 2026{\natexlab{b}}.

\bibitem[Zhao et~al.(2025)Zhao, Kwon, Streli, Pollefeys, and Holz]{zhao2025egopressure}
Yiming Zhao, Taein Kwon, Paul Streli, Marc Pollefeys, and Christian Holz.
\newblock {EgoPressure}: A dataset for hand pressure and pose estimation in egocentric vision.
\newblock In \emph{Proceedings of the Computer Vision and Pattern Recognition Conference}, 2025.

\bibitem[Zhou et~al.(2020)Zhou, Habermann, Xu, Habibie, Theobalt, and Xu]{zhou2020monocular}
Yuxiao Zhou, Marc Habermann, Weipeng Xu, Ikhsanul Habibie, Christian Theobalt, and Feng Xu.
\newblock Monocular real-time hand shape and motion capture using multi-modal data.
\newblock In \emph{Proceedings of the IEEE/CVF Conference on Computer Vision and Pattern Recognition}, pp.\  5346--5355, 2020.

\end{thebibliography}
\bibliographystyle{iclr2027_conference}

\clearpage
\appendix
\etocsettocstyle{\section*{Appendix Contents}}{\bigskip}
\etocsetnexttocdepth{subsection}
\etocsettagdepth{maintext}{none}
\etocsettagdepth{appendixtoc}{subsection}
\etocdepthtag.toc{appendixtoc}
\tableofcontents
\clearpage
\section{Implementation Details}
\label{app:impl}

\subsection{\texorpdfstring{$\boldsymbol\beta$}{beta}-cache Inference Procedure}
\label{app:beta_cache_algo}

\begin{algorithm}[h]
\caption{$\boldsymbol\beta$-cache inference for one video, run per hand}
\label{alg:beta_cache}
\begin{algorithmic}[1]
\Require clips $\mathcal{C}_{1},\dots,\mathcal{C}_{W}$; visibility threshold $T_{v}$; decoder $\mathcal{D}$
\State $\boldsymbol\beta_{\mathrm{cache}} \gets \emptyset$, \ $\mathcal{P} \gets [\,]$ \Comment{no shape yet; no clips held back}
\For{$w = 1$ \textbf{to} $W$}
  \If{$\boldsymbol\beta_{\mathrm{cache}} \neq \emptyset$}
    \State \textbf{output} $\mathcal{D}(w \mid \boldsymbol\beta_{\mathrm{cache}})$ \Comment{shape known: decode once}
  \Else
    \State $(\hat{\boldsymbol\beta}^{w}, v^{w}) \gets \mathcal{D}(w)$ \Comment{probe: shape-head output and visible fraction}
    \If{$v^{w} > T_{v}$}
      \State $\boldsymbol\beta_{\mathrm{cache}} \gets \hat{\boldsymbol\beta}^{w}$ \Comment{$\mathcal{C}_{w}$ is the first well-visible clip, \ie $w^{*}$}
      \ForAll{$u \in \mathcal{P} \cup \{w\}$}
        \State \textbf{output} $\mathcal{D}(u \mid \boldsymbol\beta_{\mathrm{cache}})$ \Comment{re-decode the held-back clips}
      \EndFor
      \State $\mathcal{P} \gets [\,]$
    \Else
      \State append $w$ to $\mathcal{P}$ \Comment{hold back until the shape is known}
    \EndIf
  \EndIf
\EndFor
\ForAll{$u \in \mathcal{P}$}
  \State \textbf{output} $\mathcal{D}(u \mid \hat{\boldsymbol\beta}^{u})$ \Comment{no clip qualified: fall back to per-clip shapes}
\EndFor
\end{algorithmic}
\end{algorithm}

\begin{figure}[t]
\centering
\includegraphics[width=\linewidth]{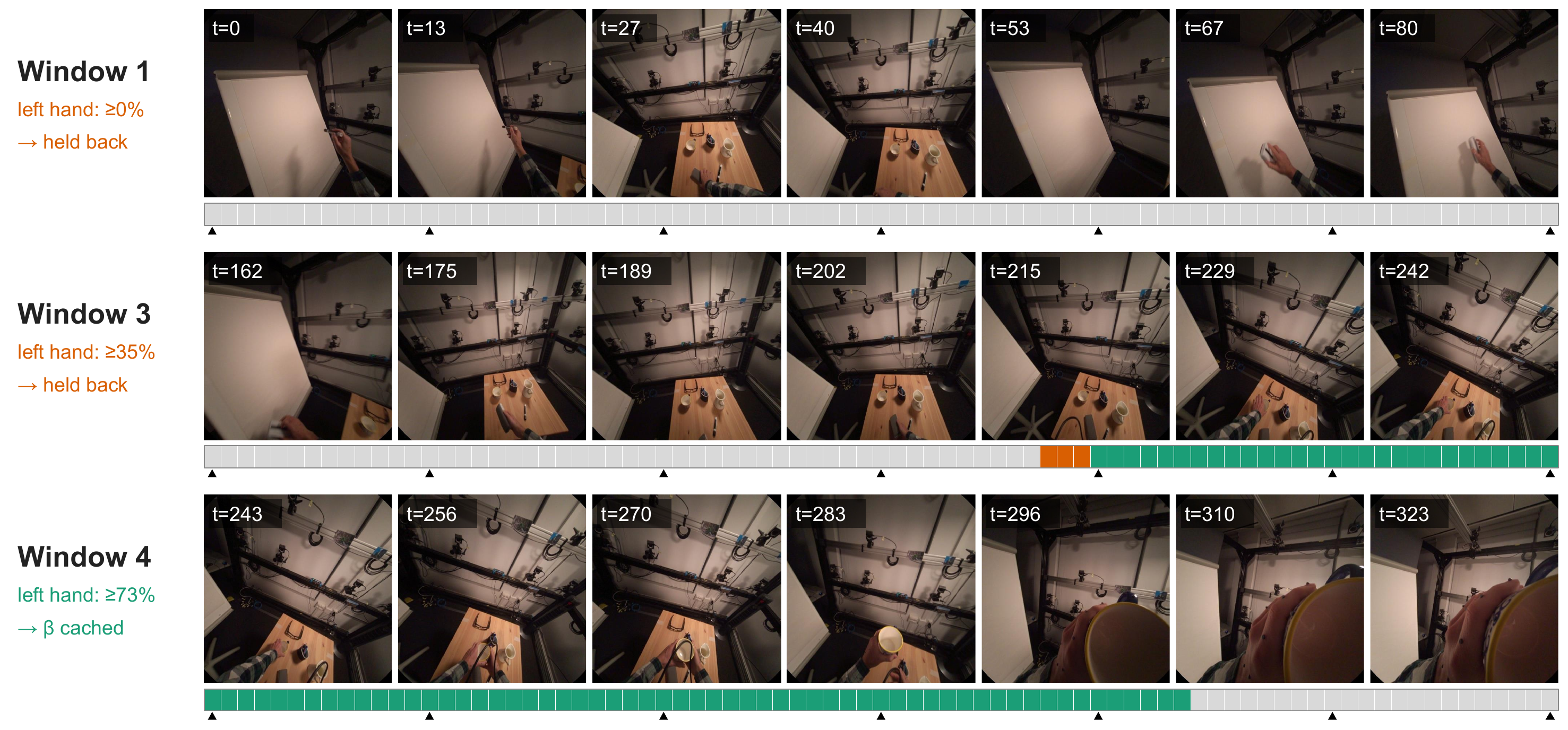}
\vspace{-0.3cm}
\caption{
\textbf{$\boldsymbol\beta$-cache inference on a test recording.}
Left hand of a HOT3D Aria test recording decoded in $81$-frame windows. Seven evenly spaced frames are shown per window, with the strip below marking frame-level \textcolor{bcpresent}{presence} or \textcolor{bcabsent}{absence} and triangles indicating the displayed frames. Windows whose visibility stays under $T_v$ are \textcolor{bcheld}{held back}. The left hand first exceeds $T_v$ in window~4, so its shape becomes $\boldsymbol\beta_{\mathrm{cache}}$ and windows~1--3 are decoded again; the right hand is cached from window~1.
}
\label{fig:beta_cache}
\end{figure}

Algorithm~\ref{alg:beta_cache} details the inference-time shape protocol of Section~\ref{method:decoder}.
Videos are decoded clip by clip until the first sufficiently visible clip $\mathcal{C}_{w^{*}}$ is found, whose shape-head output becomes $\boldsymbol\beta_{\mathrm{cache}}$ and is reused for all clips; previously probed clips are then decoded again with the cached shape.
If no clip qualifies, each clip retains its own shape estimate, corresponding to the per-clip setting in Table~\ref{tab:pose_comparison}.
With an externally calibrated shape, the algorithm starts with $\boldsymbol\beta_{\mathrm{cache}}$ initialized and decodes every clip once without probing.
Figure~\ref{fig:beta_cache} illustrates this procedure on a test recording, where the first well-visible window supplies the shape shared by the entire video.

\subsection{Pseudo-Label Generation}
\label{method:dataset}
\begin{figure}[t]
    \centering
    \includegraphics[width=\linewidth]{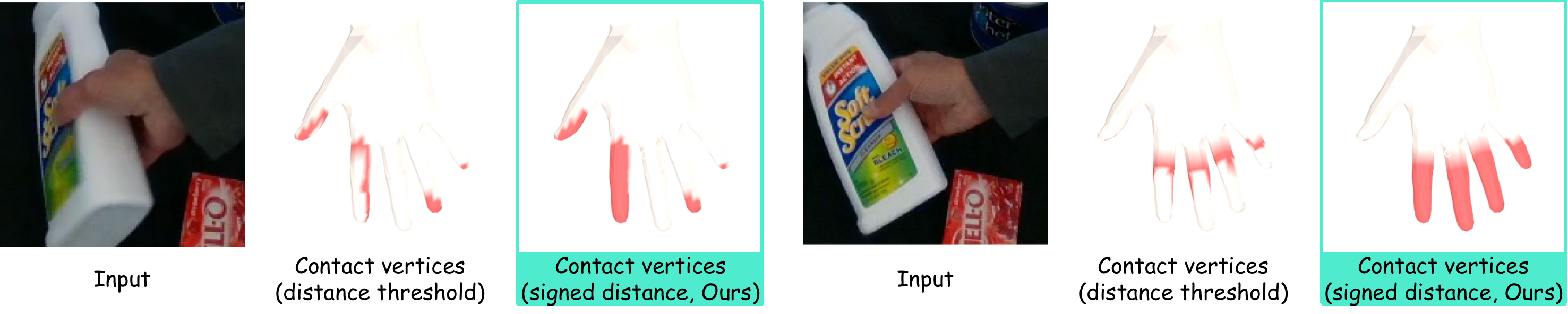}
    \caption{
    \textbf{Contact label differences between distance threshold and signed distance.}
    Contact labels from the conventional unsigned distance (middle) and our signed distance (right) on two DexYCB grasps.
    We denote the contacted vertices with the red mark.
    }
    \label{fig:signed_contact}
\end{figure}
\paragraph{Pose label conversion.}
\label{app:ik_design}
Table~\ref{tab:ik_design} compares solvers for converting the original $45$-DoF MANO labels into the constrained set $\mathcal{C}$ of Eq.~(\ref{eq:dof_constraint}) by minimizing the fitting objective in Eq.~(\ref{eq:ik_objective}).
All solvers start from the same projection onto $\mathcal{C}$ and optimize the same $20$ active finger DoF, \ie the three axes of each non-thumb MCP joint and the bend of each non-thumb PIP and DIP joint.
We evaluate the mean per-joint distance between the converted pose and the original label over the $15$ hand joints and $5$ fingertips, using the label's shape $\boldsymbol\beta$; frames come from DexYCB~\citep{chao2021dexycb} and ARCTIC~\citep{fan2023arctic}, and the cost column reports wall-clock time relative to ours on the same frames.

Adam with many steps is the most accurate solver, \eg $200$ steps reduce the median error to $0.34$\,mm on DexYCB and $0.65$\,mm on ARCTIC, but at $283\times$ our cost.
Since the conversion runs over every frame of every training set, we exclude it because such preprocessing cost would hinder further scaling of the training corpus.
Among the remaining solvers, we first discard those with unacceptable worst-case errors, \ie the undamped Gauss--Newton step, whose near-singular Jacobian at extended fingers raises the $99$th percentile on ARCTIC to $13.2$\,mm and the worst-frame error to $25.0$\,mm.
We then discard cyclic coordinate descent and the root-to-tip sequential solve, which match our error within $0.05$\,mm but cost $3\times$ and $8\times$ as much.
This leaves the damped Gauss--Newton step of Eq.~(\ref{eq:ik_step}), which we run for three iterations at $\lambda{=}10^{-3}$ with a finite-difference Jacobian.
The fit is insensitive to the damping value between $10^{-3}$ and $10^{-5}$, and retaining the fingertips in the residual reduces errors at contact-relevant vertices, \eg from $0.66$ to $0.47$\,mm on DexYCB.
Each converted label stores its residual, which is $0.5$-$1$\,mm across the corpus.

\begin{table}[t]
\centering
\small
\setlength{\tabcolsep}{4pt}
\resizebox{\columnwidth}{!}{%
\begin{tabular}{llcccccccc}
\toprule
& & & \multicolumn{3}{c}{DexYCB} & \multicolumn{3}{c}{ARCTIC} & \\
\cmidrule(lr){4-6} \cmidrule(lr){7-9}
Step rule & $\lambda$ & Iters & Median $\downarrow$ & P99 $\downarrow$ & Max $\downarrow$ & Median $\downarrow$ & P99 $\downarrow$ & Max $\downarrow$ & Cost $\downarrow$ \\
\midrule
None (projection only) & -- & -- & 1.03 & 2.34 & 3.23 & 1.98 & 3.96 & 7.85 & $0.1\times$ \\
First-order descent & -- & 3 & 1.01 & 2.32 & 3.20 & 1.95 & 3.91 & 7.80 & $1\times$ \\
Cyclic coordinate descent & -- & 3 & \second{0.42} & \best{0.77} & \second{1.43} & \second{0.91} & 1.54 & \second{4.41} & $3\times$ \\
Root-to-tip sequential & $10^{-3}$ & 3 & 0.47 & 0.80 & \second{1.43} & 0.93 & \second{1.46} & \best{4.34} & $8\times$ \\
Gauss--Newton & $0$ & 3 & \best{0.34} & 1.27 & 9.88 & \best{0.72} & 13.21 & 24.96 & $1\times$ \\
\midrule
\textbf{Ours} (damped Gauss--Newton) & $10^{-3}$ & 3 & 0.47 & \second{0.80} & \best{1.41} & 0.92 & \best{1.44} & 5.07 & $1\times$ \\
\midrule
Adam & -- & 50 & 0.37 & 0.70 & 1.46 & 0.83 & 1.36 & 4.62 & $69\times$ \\
Adam & -- & 200 & 0.34 & 0.63 & 1.15 & 0.65 & 1.16 & 2.08 & $283\times$ \\
\bottomrule
\end{tabular}%
}
\caption{\textbf{Constrained-label fitting.} Label error (mm), \ie the mean per-joint distance between the constrained pose and the original label over the $15$ hand joints and the $5$ fingertips, on 4057 DexYCB and 4089 ARCTIC frames. Every refinement starts from the same projection and runs the same number of iterations, so the rows differ only in the step rule. The undamped solve attains a competitive median but its worst frames degrade by an order of magnitude, which is the failure mode that matters when the output is a training label; damping bounds it. Cost is wall-clock time relative to ours on the same frames and hardware. We denote \bestcap{best} and \secondcap{second best} values.}
\label{tab:ik_design}
\end{table}

We convert only the MANO parameters and leave all 2D and 3D keypoint labels unchanged.
Most of our sources obtain 3D keypoints first, \eg from marker-based motion capture, multi-view triangulation, or a head-mounted device, and fit MANO to these keypoints afterwards.
Thus, the anatomically infeasible twist and spread angles in the original MANO labels arise from the unconstrained MANO fitting rather than from the keypoint annotations themselves, as the fitting process can use the extra DoF of MANO to absorb residual keypoint error.
Refitting the keypoints to the constrained MANO would moreover alter the ground truth against which every method is evaluated, reducing benchmark reproducibility and potentially tailoring the evaluation to our parameterization.
Keeping the original keypoints therefore ensures that all comparisons in Section~\ref{sec:experiments} remain against the datasets as released.

The conversion is applied to every stage-1 training source, \ie ARCTIC~\citep{fan2023arctic}, HOT3D~\citep{banerjee2024introducing}, H2O~\citep{kwon2021h2o}, OakInk2~\citep{zhan2024oakink2}, DexYCB~\citep{chao2021dexycb}, HO3D~\citep{hampali2020honnotate}, and HRDexDB~\citep{lim2026hrdexdb}.
Motion reconstruction is evaluated against the released keypoints on $210$k hand frames from the HOT3D~\citep{banerjee2024introducing} test split, $53$k from the ARCTIC~\citep{fan2023arctic} egocentric validation split, and $174$k from the EgoDex~\citep{hoque2026egodex} test split; EgoDex is unseen during training by our method and by every baseline.

\paragraph{Contact label extraction.}
We derive dense per-vertex contact labels from hand-object interaction datasets that provide ground-truth hand and object meshes.
Let $\mathbf{x}_{t,v} \in \mathbb{R}^{3}$ denote the world-space position of MANO vertex $v \in V$ at frame $t$, and let $\mathcal{S}^{\text{obj}}_{t}$ denote the posed object surface.
A natural choice is the \emph{unsigned} point-to-surface distance used by distance-based contact annotation~\citep{zhang2026human, jung2026learning}: a vertex is considered in contact when it lies within $\delta_c$ of the object surface.
However, mocap-fitted hand meshes frequently interpenetrate the object.
A vertex that penetrates deeper than $\delta_c$ can therefore be \emph{farther} than $\delta_c$ from the object surface, causing the unsigned rule to incorrectly label it as non-contact.
As Figure~\ref{fig:signed_contact} shows for two DexYCB grasps, this can erase fingertips where the grip is firmest and fragment the contact region; in the right example, the unsigned rule misses $253$ of the $364$ ground-truth contact vertices.
We therefore use the \emph{signed} distance
\begin{equation}
  d_{t,v} \;=\; \pm\,\mathrm{dist}\!\left(\mathbf{x}_{t,v},\,
  \mathcal{S}^{\text{obj}}_{t}\right),
  \label{eq:signed_distance}
\end{equation}
where the sign is negative if $\mathbf{x}_{t,v}$ lies inside the object, \ie if a pseudo-normal inside-outside test marks it as interior, and label
\begin{equation}
  c_{t,v} \;=\; \mathbf{1}\!\left[\, d_{t,v} \le \delta_c \,\right],
  \qquad \delta_c = 5\,\mathrm{mm}.
  \label{eq:contact_label}
\end{equation}

A penetrating vertex satisfies $d_{t,v} < 0 \le \delta_c$ and is therefore labeled as contact regardless of penetration depth, recovering the clean, complete contact regions shown on the right of Figure~\ref{fig:signed_contact}.
We store the continuous signed distances along with $d_{t,v}$, which enables $\delta_c$ to be tuned at training time.

These geometric labels supervise only the contact head in stage-2 and come from the sources that release object meshes, \ie ARCTIC~\citep{fan2023arctic}, HOT3D~\citep{banerjee2024introducing}, DexYCB~\citep{chao2021dexycb}, HO3D~\citep{hampali2020honnotate}, HRDexDB~\citep{lim2026hrdexdb}, and OakInk2~\citep{zhan2024oakink2}; they also serve as the contact ground truth of the DexYCB test, HOT3D test, and ARCTIC egocentric validation splits, which carry no force.

\paragraph{Force label extraction.}
For the two force-labeled sources, we follow the label revision of HOPE~\citep{jeon2026hope}, so that our training targets and the test ground truth in Table~\ref{tab:tactile_merged} use a consistent convention.
On OpenTouch~\citep{song2025opentouch}, the glove provides a $16\times16$ taxel grid with $169$ active taxels.
We lift the readings onto the MANO surface using the calibrated taxel-to-vertex correspondence $\mathcal{T}(v)$: each taxel value is assigned to every vertex it covers, while vertices covered by multiple taxels take their mean,
\begin{equation}
\bar{p}_{t,v} \;=\; \frac{1}{|\mathcal{T}(v)|}
\sum_{i \in \mathcal{T}(v)} \pi_{t,i},
\label{eq:force_lift}
\end{equation}
which covers $205$ of the $778$ vertices.
The remaining vertices are unsensed and are excluded from both the contact and force losses using a per-vertex mask.

Since the glove signal drifts and does not return to zero between grasps, HOPE gates it using a per-clip frame state.
Specifically, the total pressure is tracked against an adaptive floor with hysteresis and morphological cleaning; frames below the floor are labeled as no-contact and assigned zero force, while no-contact frames within three frames of a transition are excluded from the contact loss as ambiguous.
Training uses the $1469$ clips in the authors' allowlist, while the test split uses their human-verified frame states on the raw glove pressure, \ie the ground truth used by their evaluation code.

On PressureVisionDB~\citep{grady2022pressurevision}, we use the vertex-level pressure targets derived by the HOPE authors from the pressure pad and their MANO fits, restricted to the $421$ palm-side vertices that can contact the pad; frames without a target are left unsupervised.
Both sources use kPa as the force unit, and force-derived contact is defined as $c_{t,v} = \mathbf{1}[\,\bar{p}_{t,v} > 1\,\mathrm{kPa}\,]$ within the sensed vertex set.

OpenTouch and PressureVisionDB are the two force-labeled sources of stage-2, used together with the contact-only sources above, and tactile estimation is evaluated on their test splits as defined by HOPE.

\subsection{Ray-Based Camera Solver}
\label{app:camera}
A zero-initialized linear \emph{ray head} on the tapped feature grid $\mathbf{F}$ predicts a unit viewing ray per cell, which is averaged over the $T'$ latent frames to obtain one ray field $\hat{\mathbf{r}}\in\mathbb{R}^{h\times w\times 3}$ per clip, since the intrinsics are constant within a clip.
It is supervised by $\mathcal{L}_{\mathrm{ray}}$, the cosine distance to the calibrated pixel rays $\mathbf{r}$ defined in Appendix~\ref{app:loss}, and its Fourier encoding serves as the ray positional embedding of Section~\ref{method:decoder}.

The translation $\hat{\boldsymbol\tau}_{t} = (\hat{\tau}_{x,t},\hat{\tau}_{y,t},\hat{\tau}_{z,t})$ is recovered by the mixed PnP scheme of ACE-Ego-Hand~\citep{liu2026aceegohand}.
The depth head of the hand token directly predicts $\hat{\tau}_{z,t}$, thereby fixing the depth $z_{t,j}$ of each posed canonical joint $\mathbf{J}^{\mathrm{can}}_{t,j}$.
The in-plane components are then obtained by the closed-form least-squares solution
\begin{equation}
\begin{aligned}
(\hat{\tau}_{x,t},\hat{\tau}_{y,t}) &= \operatorname*{arg\,min}_{\tau_{x},\tau_{y}}
\sum_{j} w_{t,j}\,\Big\| \frac{\mathbf{J}^{\mathrm{can}}_{t,j,xy} + (\tau_{x},\tau_{y})}{z_{t,j}} - \mathbf{b}_{t,j} \Big\|_{2}^{2}, \\
\mathbf{b}_{t,j} &= \Pi^{-1}_{\mathbf{K}}(\hat{\mathbf{a}}_{t,j}),
\end{aligned}
\label{eq:mixed_pnp}
\end{equation}
where $\mathbf{b}_{t,j}$ denotes the bearing, \ie the normalized image coordinate obtained by back-projecting the predicted 2D anchor $\hat{\mathbf{a}}_{t,j}$ of Appendix~\ref{app:loss} through the calibrated intrinsics $\mathbf{K}$.
The weight is $w_{t,j} = z_{t,j}^{-2}$, and the solution uses only joints whose anchors lie inside the frame and whose depths are positive.
When fewer than a minimum number of joints vote, or the refit residual exceeds a threshold, the solver falls back to placing the wrist on its own bearing at depth $\hat{\tau}_{z,t}$.
The camera-frame joints are then $\widehat{\mathbf{J}}^{\mathrm{cam}}_{t} = \mathbf{J}^{\mathrm{can}}_{t} + \hat{\boldsymbol\tau}_{t}$, which are used in the 3D, 2D, and temporal losses.

All reported models use the calibrated $\mathbf{K}$ of each dataset; the ray head is used only for the positional embedding and auxiliary ray supervision.

\subsection{Objective Function and Evaluation Metrics}
\label{app:loss}
\label{app:metrics}
This section gives the closed-form definitions of all training losses in Section~\ref{method:training} and evaluation metrics in Section~\ref{sec:experiments}.
Throughout, $\langle\cdot\rangle$ denotes the mean over the valid entries of a mask.
For training losses, $m_{t,s}\in\{0,1\}$ indicates the availability of the corresponding supervision type, \eg MANO parameters, 3D keypoints, or 2D keypoints, for frame $t$ and hand $s$, allowing heterogeneous sources to be trained jointly in a single mixture.
Here, $t$ indexes frames, $s\in\{\mathrm{L},\mathrm{R}\}$ indexes hands, and $j$ indexes the $J{=}21$ joints; hats denote predictions.

\paragraph{Rotations and shape.}
Let $\widehat{\mathrm{R}}_{t,j}$ be the assembled rotation stack of Section~\ref{method:decoder}, \ie the global orientation and the $15$ joint rotations, and $\mathrm{R}_{t,j}$ the ground truth. The rotation and shape losses are:
\begin{equation}
  \mathcal{L}_{\mathrm{rot}} = \big\langle \arccos\tfrac{\mathrm{tr}(\widehat{\mathrm{R}}^{\top}\mathrm{R})-1}{2} \big\rangle
  + \big\langle \|\widehat{\mathrm{R}}-\mathrm{R}\|_{F}^{2} \big\rangle,
  \qquad
  \mathcal{L}_{\beta} = \big\langle \|\widehat{\boldsymbol\beta}-\boldsymbol\beta\|_{1} \big\rangle,
\end{equation}
where the shape loss is applied only to hands with ground-truth $\boldsymbol\beta$ that are not teacher-forced in that step, as described in Section~\ref{method:decoder}, ensuring that the shape head is supervised under the same unconditioned setting used at test time.

\paragraph{3D joints.}
The wrist-relative term supervises articulation for both the MANO-posed joints and the joints directly regressed by the joint tokens, while the camera-frame and wrist terms supervise absolute placement through the recovered translation:
\begin{equation}
  \mathcal{L}_{\mathrm{3D}} =
  \lambda_{\mathrm{rel}}\big\langle \|\widehat{\mathbf{J}}^{\mathrm{rel}}-\mathbf{J}^{\mathrm{rel}}\|_{1} \big\rangle
  + \lambda_{\mathrm{cam}}\big\langle \|\widehat{\mathbf{J}}^{\mathrm{cam}}-\mathbf{J}^{\mathrm{cam}}\|_{1} \big\rangle
  + \lambda_{\mathrm{w}}\big\langle \|\widehat{\mathbf{J}}^{\mathrm{cam}}_{\mathrm{wrist}}-\mathbf{J}^{\mathrm{cam}}_{\mathrm{wrist}}\|_{1} \big\rangle.
\end{equation}

\paragraph{2D terms.}
The soft-argmax anchors and the reprojection of the posed 3D joints are supervised in normalized image coordinates:
\begin{equation}
  \mathcal{L}_{\mathrm{2D}} =
  \big\langle \|\hat{\mathbf{a}}-\mathbf{u}\|_{1} \big\rangle_{\mathrm{in}}
  + \big\langle \|\Pi(\widehat{\mathbf{J}}^{\mathrm{cam}})-\mathbf{u}\|_{1} \big\rangle_{\mathrm{in,front}}
  + \tfrac{1}{2}\big\langle \|\Pi(\widehat{\mathbf{J}}^{\mathrm{cam}}_{\mathrm{wrist}})-\mathbf{u}_{\mathrm{wrist}}\|_{1} \big\rangle_{\mathrm{in,front}},
\end{equation}
where $\langle\cdot\rangle_{\mathrm{in}}$ masks to confidence-weighted targets whose projections lie inside the frame, \ie the valid targets for the soft-argmax, and $\langle\cdot\rangle_{\mathrm{front}}$ additionally requires the predicted joint to lie in front of the camera.

\paragraph{Translation, presence, smoothness, rays.}
The translation target is the offset between the ground-truth wrist and the wrist of the ground-truth-posed canonical MANO, $\boldsymbol\tau^{\mathrm{gt}}_{t} = \mathbf{J}^{\mathrm{cam}}_{\mathrm{wrist},t} - \mathbf{J}^{\mathrm{can}}_{\mathrm{wrist},t}$, supervised with an $\mathcal{L}_1$ loss;
Existence and visibility logits are supervised with binary cross-entropy on frames with known presence;
$\mathcal{L}_{\mathrm{tmp}}$ penalizes the second temporal difference, \ie the acceleration, of the camera-frame joints over valid frames;
Finally, $\mathcal{L}_{\mathrm{ray}} = \langle 1 - \hat{\mathbf{r}}^{\top}\mathbf{r}\rangle$ is the mean cosine distance between the predicted and calibrated pixel rays over the token grid.

\paragraph{Detection gate.}
All ``-p'' pose metrics follow the detection-gated protocol of ACE-Ego-Hand~\citep{liu2026aceegohand}.
Let $o_{t,s}\in\{0,1\}$ denote whether the ground-truth hand is on screen, $\hat{p}_{t,s}$ the predicted presence probability, $\mathcal{B}(\cdot)$ the axis-aligned bounding box of a set of image points, and $\mathcal{B}_{1.1}(\mathbf{u}_{t,s})$ the ground-truth keypoint box dilated by $10\%$.
A prediction counts as a detection when all three conditions below hold:
\begin{equation}
\begin{aligned}
  d_{t,s} = {}& \mathbb{1}\big[\hat{p}_{t,s} > 0.5\big]\;
            \mathbb{1}\big[\Pi(\widehat{\mathbf{J}}^{\mathrm{cam}}_{t,s})\ \text{has a joint inside the frame with}\ \hat{z}>0\big] \\
            & \cdot\;\mathbb{1}\big[\mathcal{B}(\Pi(\widehat{\mathbf{J}}^{\mathrm{cam}}_{t,s})) \cap \mathcal{B}_{1.1}(\mathbf{u}_{t,s}) \neq \emptyset\big].
\end{aligned}
\end{equation}
Matched frames, \ie $d_{t,s}=1$, contribute their actual errors, while a missed on-screen hand incurs a fixed \emph{canonical-MANO penalty}:
The error of the wrist-aligned rest pose with the mean shape for the 3D terms and the image diagonal for the 2D term.
This prevents the metrics from being improved by dropping difficult frames.

\paragraph{F$_{\mathrm{Acc}}$.}
Frame accuracy is the fraction of frames with neither a missed on-screen hand nor a spurious detection:
\begin{equation}
  \mathrm{F}_{\mathrm{Acc}} = \frac{1}{T}\sum_{t=1}^{T}\mathbb{1}\Big[\sum_{s} o_{t,s}\,(1-d_{t,s}) + (1-o_{t,s})\,f_{t,s} = 0\Big],
\end{equation}
where $f_{t,s}=1$ indicates a hand predicted to be present inside the frame despite a negative annotation.
On bimanual datasets without explicit absence labels, only the first term is active.

\paragraph{MPJPE-p and PA-p.}
Let $\bar{\mathbf{J}}_{t,s} = \mathbf{J}_{t,s} - \mathbf{J}_{t,s,\mathrm{wrist}}$ denote the wrist-aligned joints and $e^{\mathrm{can}}_{t,s}$ the canonical-MANO penalty of a missed hand.
MPJPE-p is the mean per-joint error in mm over the on-screen ground-truth hands, where a detected hand contributes its wrist-aligned error and a missed hand contributes the penalty:
\begin{equation}
  \mathrm{MPJPE\text{-}p} = \Big\langle d_{t,s}\,\tfrac{1}{J}\textstyle\sum_{j}\big\|\bar{\widehat{\mathbf{J}}}_{t,s,j}-\bar{\mathbf{J}}_{t,s,j}\big\|_{2} + (1-d_{t,s})\,e^{\mathrm{can}}_{t,s} \Big\rangle_{o_{t,s}=1}.
\end{equation}
PA-p is defined in the same way after aligning $\bar{\widehat{\mathbf{J}}}_{t,s}$ to $\bar{\mathbf{J}}_{t,s}$ with a Procrustes fit, \ie the optimal rotation, translation, and scale.
The penalty of a missed hand is aligned in the same way.

\paragraph{EPE2D-p.}
Let $\mathcal{V}_{t,s}$ be the ground-truth joints inside the frame and $(W,H)$ the image size. EPE2D-p is the mean 2D joint error in pixels of the evaluation resolution over the on-screen ground-truth hands, where a missed hand contributes the image diagonal:
\begin{equation}
  \mathrm{EPE2D\text{-}p} = \Big\langle d_{t,s}\,\tfrac{1}{|\mathcal{V}_{t,s}|}\textstyle\sum_{j\in\mathcal{V}_{t,s}}\big\|\Pi(\widehat{\mathbf{J}}^{\mathrm{cam}}_{t,s,j})-\mathbf{u}_{t,s,j}\big\|_{2} + (1-d_{t,s})\sqrt{W^{2}+H^{2}} \Big\rangle_{o_{t,s}=1}.
\end{equation}

\paragraph{MPJPE$^{+\mathrm{OOS}}$.}
This metric bypasses the detection gate and measures the wrist-aligned error over \emph{all} frames with a ground-truth 3D hand $g_{t,s}=1$, including out-of-sight ones:
\begin{equation}
  \mathrm{MPJPE}^{+\mathrm{OOS}} = \Big\langle \tfrac{1}{J}\textstyle\sum_{j}\big\|\bar{\widehat{\mathbf{J}}}_{t,s,j}-\bar{\mathbf{J}}_{t,s,j}\big\|_{2} \Big\rangle_{g_{t,s}=1}.
\end{equation}
This metric therefore evaluates temporal extrapolation across periods of missing observations.

\paragraph{Jitter.}
Jitter measures temporal consistency rather than accuracy.
Let $\mathcal{R}$ be the set of maximal runs of at least three consecutive frames in which the hand is predicted present and is on screen in the ground truth.
Jitter is the joint-averaged norm of the second temporal difference of the camera-frame joints in mm/frame$^{2}$, averaged first within each run and then across runs:
\begin{equation}
  \mathrm{Jitter} = \frac{1}{|\mathcal{R}|}\sum_{r\in\mathcal{R}}\ \frac{1}{|r|-2}\sum_{t\in r}\ \frac{1}{J}\sum_{j}\big\|\widehat{\mathbf{J}}^{\mathrm{cam}}_{t+1,j} - 2\widehat{\mathbf{J}}^{\mathrm{cam}}_{t,j} + \widehat{\mathbf{J}}^{\mathrm{cam}}_{t-1,j}\big\|_{2}.
\end{equation}
The 2D detection gate used for the accuracy metrics is not applied, so that each tracker is evaluated on all of its predictions rather than only on well-localized ones.
Since the evaluated frame sets differ across methods, we additionally verified that restricting all methods to the frames detected by every method leaves the ranking unchanged on all three test sets.

\paragraph{Shape jitter.}
$\sigma_{\mathrm{shape}}$ quantifies the within-recording drift in predicted hand \emph{size}, despite the subject's hand remaining consistent.
Let $\ell(\boldsymbol\beta)$ be the length of the middle-finger bone chain, \ie wrist-MCP-PIP-DIP-tip, of the MANO template posed with shape $\boldsymbol\beta$ at the identity pose $\boldsymbol\theta_{0}$. The bone length and the resulting shape jitter are:
\begin{equation}
  \ell(\boldsymbol\beta) = \sum_{k=1}^{4}\big\|\mathbf{J}_{k}(\boldsymbol\beta,\boldsymbol\theta_{0}) - \mathbf{J}_{k-1}(\boldsymbol\beta,\boldsymbol\theta_{0})\big\|_{2},
  \qquad
  \sigma_{\mathrm{shape}} = \Big\langle \operatorname{std}_{t\in\mathcal{T}_{c,s}}\ \ell(\hat{\boldsymbol\beta}_{t,s}) \Big\rangle_{(c,s)},
\end{equation}
where $\mathcal{T}_{c,s}$ is the set of frames of recording $c$ and hand $s$ for which the method predicts the hand as present while the ground-truth hand is on screen, and the outer mean, in mm, is taken over all recordings and hands with $|\mathcal{T}_{c,s}|\geq 30$.
Because $\ell$ is evaluated at a fixed pose, it depends only on $\hat{\boldsymbol\beta}$, so a method that uses a single shape per recording has $\sigma_{\mathrm{shape}}=0$.
Per-frame trackers re-estimate $\hat{\boldsymbol\beta}$ for every crop, causing hand size to fluctuate by several millimeters within a recording, while clip-level models re-estimate it once per window.
The $\boldsymbol\beta$-cache protocol applies the shape cached from the first well-visible window to every window in the recording, making $\sigma_{\mathrm{shape}}=0$ by construction.

\paragraph{Contact F1 and AUROC.}
Let $c_{t,v}\in\{0,1\}$ be the ground-truth contact of MANO vertex $v$ in frame $t$ and $\hat{c}_{t,v}\in[0,1]$ the predicted probability. Both scores are micro-averaged over all $|V|$ vertices of every evaluated frame:
\begin{equation}
\begin{aligned}
  \mathrm{F1} &= \frac{2\,\mathrm{TP}}{2\,\mathrm{TP}+\mathrm{FP}+\mathrm{FN}}\quad\text{with}\quad \hat{c}_{t,v}>\tau_{c}, \\
  \mathrm{AUROC} &= \Pr\big[\hat{c}_{t,v} > \hat{c}_{t',v'}\ \big|\ c_{t,v}=1,\ c_{t',v'}=0\big],
\end{aligned}
\end{equation}
where F1 is reported at the fixed operating point $\tau_{c}=0.8$, selected on the validation splits, whereas AUROC evaluates the predicted ranking independently of this threshold.
AUROC is computed from per-bin score histograms with $20$k bins, agreeing with the exact rank-based statistic to within $10^{-6}$ without materializing the full score vector.

\paragraph{Force MAE and RMSE.}
Let $f_{t,v}$ be the measured and $\hat{f}_{t,v}$ the predicted per-vertex pressure in kPa.
Over all vertices of frames with force ground truth, the two force metrics are:
\begin{equation}
  \mathrm{MAE} = \big\langle |\hat{f}_{t,v}-f_{t,v}| \big\rangle,
  \qquad
  \mathrm{RMSE} = \sqrt{\big\langle (\hat{f}_{t,v}-f_{t,v})^{2} \big\rangle}.
\end{equation}

\paragraph{Tactile ground truth and evaluation protocol.}
Contact ground truth differs by source, which matters when comparing the columns of Table~\ref{tab:tactile_merged}.
OpenTouch~\citep{song2025opentouch} labels a vertex from measured glove pressure above $1$\,kPa, whereas DexYCB~\citep{chao2021dexycb}, HOT3D~\citep{banerjee2024introducing}, and ARCTIC~\citep{fan2023arctic} derive contact geometrically as a MANO vertex lying within $5$\,mm of the object mesh.
The latter definition is also used to train HACO~\citep{jung2026learning}, making these three splits directly aligned with its training labels.
For HOT3D, the Aria and Quest recordings are evaluated jointly by summing confusion counts for F1 and score histograms for AUROC, rather than averaging the per-rig metrics.
ARCTIC provides no test-split ground truth, so we report its validation split.
We re-run HOPE~\citep{jeon2026hope} and HACO from their official code and checkpoints as described in Appendix~\ref{app:baselines}, using exactly the frames evaluated in our results, rather than quoting published numbers.

\paragraph{Q-err.}
Let $\mathbf{q}_{t}\in\mathbb{R}^{D}$ be the joint command produced by a retargeting solver at frame $t$ from the predicted hand, and let $\mathbf{q}^{\mathrm{oracle}}_{t}$ denote the command produced by the \emph{same} solver at the same frame using the \emph{ground-truth human joints}.
All retargeting metrics are computed over contiguous runs of at least three ground-truth-labeled frames, ensuring that temporal differences do not span label gaps.
Q-err is the mean absolute joint-angle deviation in degrees between the two commands:
\begin{equation}
  \text{Q-err} = \Big\langle \tfrac{1}{D}\big\|\mathbf{q}_{t}-\mathbf{q}^{\mathrm{oracle}}_{t}\big\|_{1} \Big\rangle_{t}.
\end{equation}

\paragraph{Jerk.}
Jerk is the joint-averaged absolute second temporal difference of the command, reported in $^{\circ}$/frame$^{2}$ as its \emph{median} over frames:
\begin{equation}
  \mathrm{Jerk} = \operatorname{median}_{t}\ \tfrac{1}{D}\big\|\mathbf{q}_{t+1} - 2\mathbf{q}_{t} + \mathbf{q}_{t-1}\big\|_{1}.
\end{equation}
Since the ground-truth annotations contain sparse frame-to-frame steps, we use the median instead of the mean.
The mean is dominated by these steps in the Oracle row, whereas the median better reflects the frame-to-frame noise introduced by the tracker.
We disable low-pass filter of DexPilot, so Jerk reflects the tracker rather than the filter.

\subsection{Training and Evaluation}
\label{app:exp_details}
\begin{table}[t]
\centering
\small
\setlength{\tabcolsep}{6pt}
\begin{tabular}{llcc}
\toprule
Term of Eq.~(\ref{eq:stage1_loss}) & Component & Weight & Value \\
\midrule
\multirow{2}{*}{$\mathcal{L}_{\mathrm{rot}}$} & geodesic rotation error & $\lambda_{\mathrm{geo}}$ & 1 \\
 & Frobenius rotation error & $\lambda_{\mathrm{mse}}$ & 1 \\
\midrule
$\mathcal{L}_{\beta}$ & shape & $\lambda_{\beta}$ & 0.1 \\
\midrule
\multirow{3}{*}{$\mathcal{L}_{\mathrm{3D}}$} & wrist-relative joints & $\lambda_{\mathrm{rel}}$ & 10 \\
 & camera-frame joints & $\lambda_{\mathrm{cam}}$ & 5 \\
 & camera-frame wrist & $\lambda_{\mathrm{w}}$ & 2 \\
\midrule
\multirow{3}{*}{$\mathcal{L}_{\mathrm{2D}}$} & soft-argmax anchors & $\lambda_{\mathrm{anch}}$ & 1 \\
 & joint reprojection & $\lambda_{\mathrm{reproj}}$ & 1 \\
 & wrist reprojection & $\lambda_{\mathrm{reproj,w}}$ & 0.5 \\
\midrule
$\mathcal{L}_{\tau}$ & translation & $\lambda_{\tau}$ & 1 \\
\midrule
\multirow{2}{*}{$\mathcal{L}_{\mathrm{pres}}$} & existence & $\lambda_{\mathrm{exist}}$ & 0.5 \\
 & visibility & $\lambda_{\mathrm{vis}}$ & 0.25 \\
\midrule
$\mathcal{L}_{\mathrm{tmp}}$ & temporal smoothness & $\lambda_{\mathrm{tmp}}$ & 0.5 \\
\midrule
$\mathcal{L}_{\mathrm{ray}}$ & pixel rays & $\lambda_{\mathrm{ray}}$ & 1 \\
\bottomrule
\end{tabular}
\caption{Stage-1 loss weights. Each row is one component of the corresponding term of Eq.~(\ref{eq:stage1_loss}).}
\label{tab:loss_weights}
\end{table}

\paragraph{Training.}
Training videos are sampled as clips of $T{=}81$ frames at an input height of $480$.
The RLHND decoder uses width $D{=}384$, $4$ layers, $8$ heads, and $4$ register tokens, with LoRA at rank $64$.
Both stages use AdamW with $\beta_1{=}0.9$, $\beta_2{=}0.95$, weight decay $0.01$, $200$ warmup steps followed by cosine decay, gradient clipping at $1.0$, and bf16 mixed precision. Learning rates are $2{\times}10^{-4}$ for the decoder, ray head, and tactile expert, $1{\times}10^{-4}$ for LoRA, and $2{\times}10^{-5}$ for the patch embedding.
Stage-1 trains for $20$k steps on $8$ H100 GPUs with $2$ clips per GPU and $2$-step gradient accumulation, giving an effective batch size of $32$. Stage-2 trains the tactile expert for $20$k steps on a single A100 with a batch size of $2$, while freezing the entire stage-1 model so that only the expert and vertex readout receive gradients.
The stage-1 loss weights in Eq.~(\ref{eq:stage1_loss}) are listed in Table~\ref{tab:loss_weights}. Stage-2 uses $\lambda_{c}=\lambda_{F}=\lambda_{\pi}=1$ with a positive-class weight of $10$ in $\mathrm{BCE}_{w}$.

\paragraph{Evaluation.}
\label{app:baselines}
We run every baseline with its official code and released weights on exactly the frames we score, rather than quoting published numbers. This ensures that all methods are evaluated on the same frames.
The crop-based per-frame trackers HaMeR~\citep{pavlakos2024reconstructing}, WiLoR~\citep{potamias2024wilor}, HaWoR~\citep{zhang2025hawor}, and HaPTIC~\citep{ye2025haptic} share the hand detector released with WiLoR and the detection logic introduced in HaWoR.
The detector selects the highest-confidence box per side and mirrors left hands to the right-hand model. Their F$_{\mathrm{Acc}}$ therefore coincides, and their ``-p'' metrics use the same detection front-end.
HandFlow~\citep{xu2026handflow} is run through its released pipeline.
Since it has no 2D head, EPE2D-p uses its reprojected joints, and camera translation is obtained from its crop-based weak-perspective lift.
ACE-Ego-Hand~\citep{liu2026aceegohand} is run with its official inference code and checkpoint.
We use one forward pass per test window at the protocol resolution with the released MANO decode and camera solver.

For tactile estimation, HACO~\citep{jung2026learning} is run from its public release. HOPE~\citep{jeon2026hope} uses the official code and checkpoint shared by its authors on request.
Retargeting uses the same DexPilot~\citep{handa2020dexpilot} solver for every method, as in Section~\ref{sec:retargeting}.

The two RLHND rows in Table~\ref{tab:pose_comparison} share one checkpoint and differ only in the shape protocol.
One uses per-window prediction, while the other uses the $\boldsymbol\beta$-cache of Section~\ref{method:decoder}, which applies the shape cached from the first well-visible window to every window of the recording.

\begin{figure}[t]
\centering
\includegraphics[width=\linewidth]{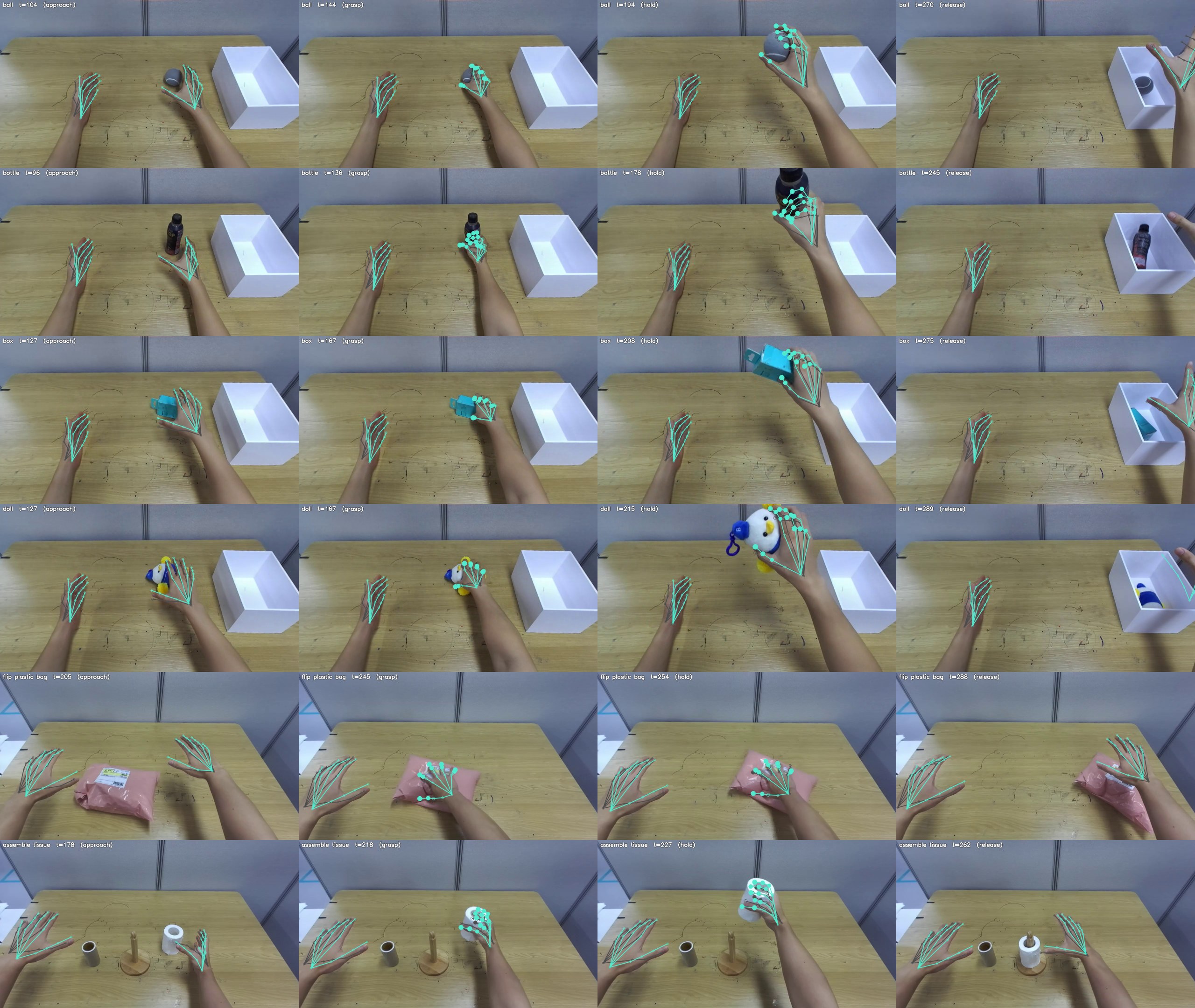}\\[2pt]
\includegraphics[width=\linewidth]{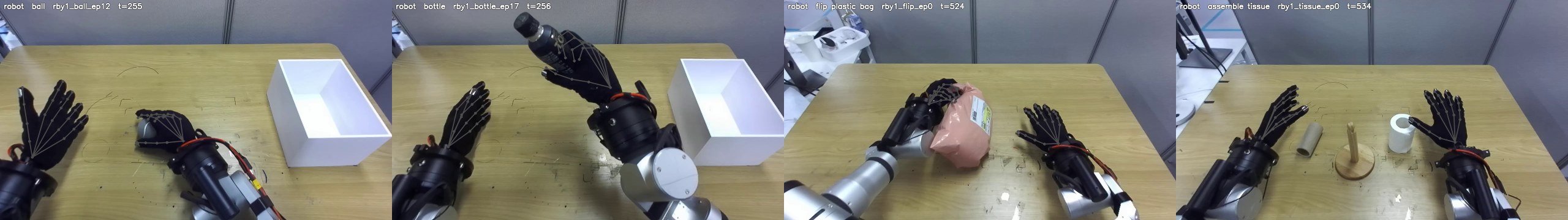}
\caption{
\textbf{Examples of the DPP training data.}
\emph{Top}: human demonstrations for the six tasks, \eg ball, bottle, box, bird, plastic bag, and tissue, across approach, grasp, hold, and release phases, viewed from the shared ego camera.
\textcolor{hawgray}{Gray} and \textcolor{oursteal}{teal} denote the original DPP, \ie \textcolor{hawgray}{HaWoR}, and \textcolor{oursteal}{RLHND} hand keypoints, respectively.
Marker size and fill indicate the corresponding contact labels.
\emph{Bottom}: teleoperated robot episodes with forward-kinematics keypoints in \textcolor{hawgray}{gray}.
}

\label{fig:dpp_dataset_examples}
\end{figure}

\begin{figure}[t]
\centering
\includegraphics[width=\linewidth]{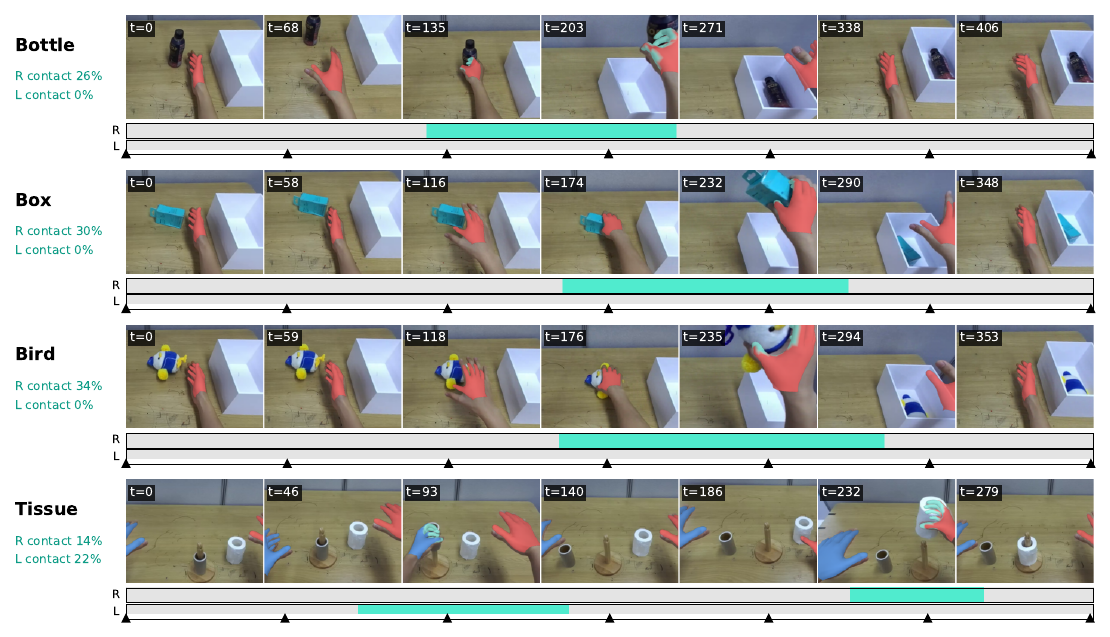}
\caption{
\textbf{Automatic contact labels from RLHND.}
We denote vertices whose predicted contact probability exceeds $0.5$ as \textcolor{oursteal}{teal}.
The strips below each row show the frame-level contact labels for the right (R) and left (L) hands, obtained by thresholding the mean contact probability over the fingertip vertices at $0.5$.
Triangles mark the displayed frames and percentages indicate the fraction of frames labeled as contact.
}
\label{fig:dpp_contact_strip}
\end{figure}

\paragraph{Training DPP.}
We explain the detailed settings for training DPP as following:

\emph{Data.}
The $200$ human demonstrations comprise $25$ demonstrations for each pick-and-place object, \eg ball, bottle, box, and bird, together with $50$ plastic-bag flips and $50$ tissue assemblies.
They are recorded at $1280\times720$ and $30$ fps using the ego rig.
The $200$ teleoperated robot episodes follow the same task split.
The 100 : 100 mixture in Table~\ref{tab:robot_learning} uses all $200$ human and all $200$ robot episodes.
Without rollout, we additionally train 100 : 10 and 100 : 50 mixtures, which keep all human episodes and sample $20$ or $100$ robot episodes using a task-balanced even stride.
Figure~\ref{fig:dpp_dataset_examples} shows example frames from both data sources.

\emph{Human hand labels.}
RLHND is run on the right ego view at the native resolution using the rectified ZED Mini intrinsics, closely following the original DPP setting for a fair comparison.
It returns the $21$ keypoints of both hands in the camera frame and in the MANO order expected by DPP, together with a per-frame validity flag from its presence head.
The original DPP labels are based on HaWoR keypoints, rescaled per frame so that the wrist aligns with the stereo depth using a $+2$ cm offset and a temporal median over $5$ frames, with the scale clipped to $[0.5,2]$.
The resulting keypoints are cleaned to remove duplicate slots, spikes, and track discontinuities.
The RLHND keypoints undergo the same alignment and cleaning before packing, ensuring that the comparison focuses on the tracker rather than the depth scale.

\emph{Contact labels.}
Since HaWoR has no contact head, both rows of Table~\ref{tab:robot_learning} use the contact labels from contact head of RLHND.
Specifically, its per-vertex contact probabilities on the $|V|$ MANO vertices are converted by the DPP packer into $21$ per-keypoint labels using a fixed vertex-to-keypoint weight map.
Figure~\ref{fig:dpp_contact_strip} shows these automatic labels on example demonstrations.
In short, The two rows differ only in the hand keypoints.

\emph{Packing and objective.}
Episodes are resampled to $20$ fps, and each observation is represented in the camera frame of its anchor step relative to a per-episode origin.
The observation consists of $18$ tokens of width $512$, comprising up to four objects represented by $512$ stereo points each, together with MiniLM embeddings of the instruction and object names, the $42$ hand keypoints, and the $9$-D camera pose.
Object points are obtained from Fast-FoundationStereo~\citep{wen2025fastfoundationstereo} depth and SAM~3~\citep{carion2025sam3} masks.
The action is a $16$-step chunk, \ie $0.8$s, of keypoint displacements, normalized per step and channel using the statistics of the DPP pre-training corpus and clipped to $\pm5$.
The objective is rectified flow, implemented as an MSE on the velocity between Gaussian noise and the target, masked when a hand is not tracked, together with a BCE loss on the $42$ contact logits computed from detached features so that the contact loss does not backpropagate into the trunk.

\emph{Optimization.}
Every policy starts from the same DiT-B checkpoint pretrained for $100$k steps and is fine-tuned for $60$k steps with AdamW using a learning rate of $10^{-4}$, weight decay of $10^{-4}$, $1$k warm-up steps followed by a constant learning rate, gradient clipping at $1.0$, batch size $128$, dropout $0.1$, and bf16 autocast on a single A100, which takes $60$-$90$ minutes per training each policy.
Snapshots are saved every $10$k steps, and the $60$k snapshot is used for deployment with $6$ Euler steps from Gaussian noise.

\begin{figure}[t]
\centering
\includegraphics[width=\linewidth]{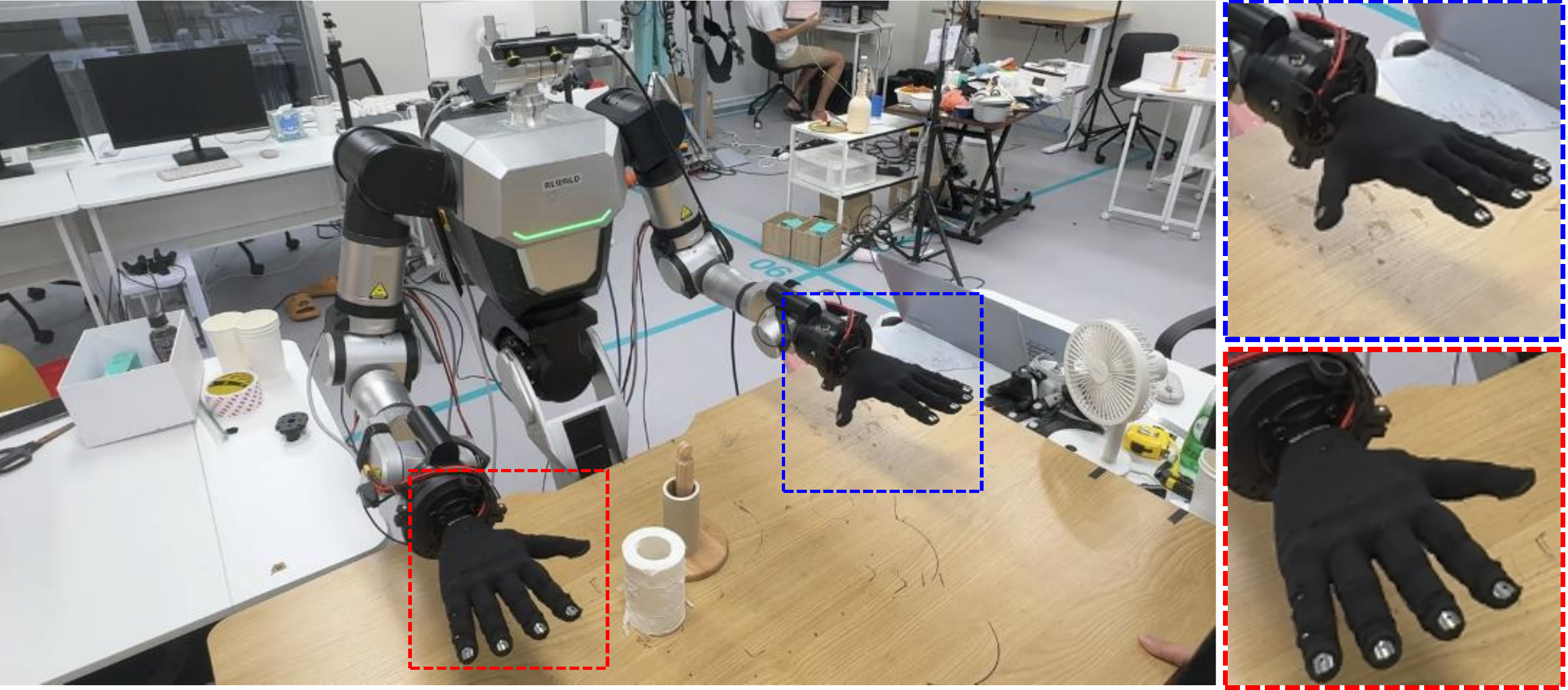}
\caption{
\textbf{Real-robot platform.} An RB-Y1 bimanual robot with a WUJI dexterous hand on each arm in the tabletop workspace used for the real-robot experiments.
The insets on the right show close-ups of the two hands, with the robot's left and right hands outlined in \textcolor{blue}{blue} and \textcolor{red}{red}, respectively.
}
\label{fig:rby1}
\end{figure}

\paragraph{Evaluation on real-robot.}
Figure~\ref{fig:rby1} shows the hardware platform.
We use an RB-Y1 bimanual robot with a WUJI Hand~2 on each arm, \ie a $7$-DoF arm and a $20$-DoF hand per side, and the policy observes the scene through the head-mounted ZED Mini stereo camera at $1280\times720$ and $30$ fps.
Human demonstrations are recorded with an identical camera rig aimed to reproduce the ego view of the robot, therefore human and robot episodes share the same image format, intrinsics, and viewpoint.
We collect $200$ human demonstrations and $200$ teleoperated robot episodes across six tasks, four single-handed pick-and-place tasks and two bimanual tasks:

\begin{itemize}[leftmargin=1.5em,itemsep=2pt]
\item \textbf{Pick and place (bottle, box, ball, bird).} The right hand picks up the object from the table and places it in the white container. The four tasks share the same procedure and differ only in the object, including a rigid bottle and box, a small ball, and a soft plush bird.
\item \textbf{Plastic bag.} The left hand picks up the plastic bag, flips it so that the barcode faces up, and hands it to the right hand, which then passes it to the other rail. This task models a conveyor sorting scenario in a logistics line, where packages must be oriented for scanning and routed to the appropriate lane.
\item \textbf{Tissue assembly.} The left hand removes the used tissue core from the holder, while the right hand picks up a new tissue roll and places it on the holder.
\end{itemize}

Both rows of the table use the same policy and training recipe, \ie a Dexterous Point Policy with DiT structure with $12$ layers of width $768$, a $16$-step action horizon, and up to four objects of $512$ points each, initialized from a checkpoint pre-trained for $100$k steps on the DPP human-video corpus and fine-tuned for $60$k steps with batch size $128$ on the mixed set.
The only difference between the rows is the \emph{hand keypoint labels} of the $200$ human episodes.
The original tracker, HaWoR, provides the keypoints for DPP, while RLHND provides them for DPP + RLHND; both rows use the contact labels of RLHND, since HaWoR does not predict contact.
The robot episodes, objects, instructions, and hyperparameters are identical.

At the inference, object points are extracted online from the stereo pair using Fast-FoundationStereo~\citep{wen2025fastfoundationstereo,wen2025foundationstereo} depth and SAM~3~\citep{carion2025sam3} masks at $5$\,Hz, and the policy runs with $6$ flow steps to predict a $16$-step chunk of keypoint displacements and per-keypoint contact probabilities for both hands.
The predicted keypoints are retargeted to the robot by inverse kinematics over the arm and hand joints with the finger abduction joints locked.
Keypoints whose contact probability exceeds $0.05$ are pushed $15$\,mm along the outward normal of their link to inject contact force.
Each task is evaluated over $16$ trials per condition with randomized object placement.

\section{Additional Experimental Results}
\label{app:results}
\subsection{Real-Robot Experiments with DPP}
\label{app:robot_results}

\begin{figure}[t]
\centering
\includegraphics[width=\linewidth]{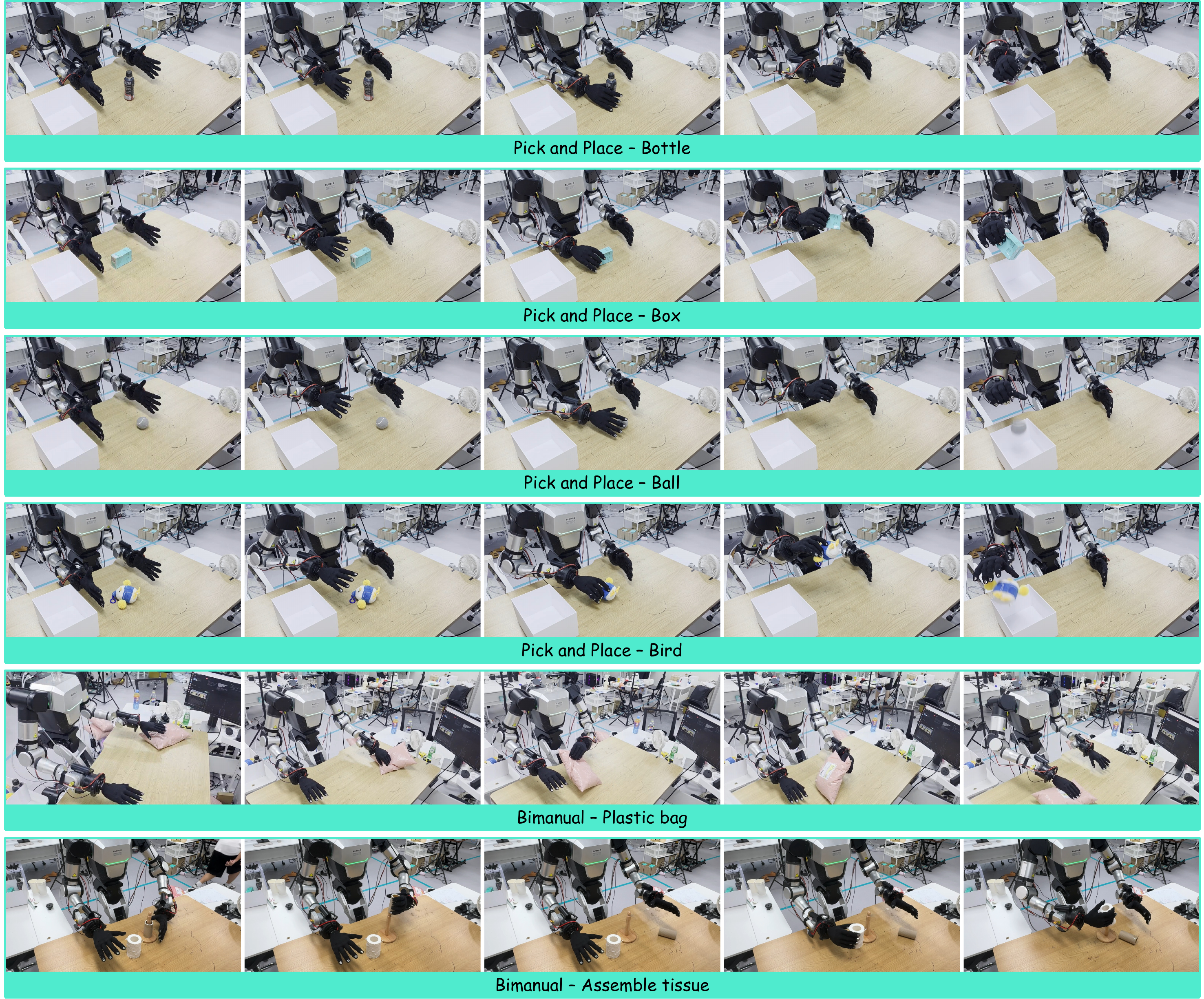}
\caption{\textbf{Real-robot rollouts of DPP + RLHND.} Five snapshots of one rollout per task of the DPP + RLHND policy on the RB-Y1 with WUJI hands, from a third-person view. The four pick-and-place tasks (bottle, box, ball, bird) have the right hand grasp the object and drop it into the white container, and the two bimanual tasks are the ones where the left hand picks up and flips the plastic bag before the right hand passes it on, and the left hand removes the used tissue core before the right hand places the new roll on the holder.}
\label{fig:robot_demo}
\end{figure}

\begin{figure}[t]
\centering
\includegraphics[width=\linewidth]{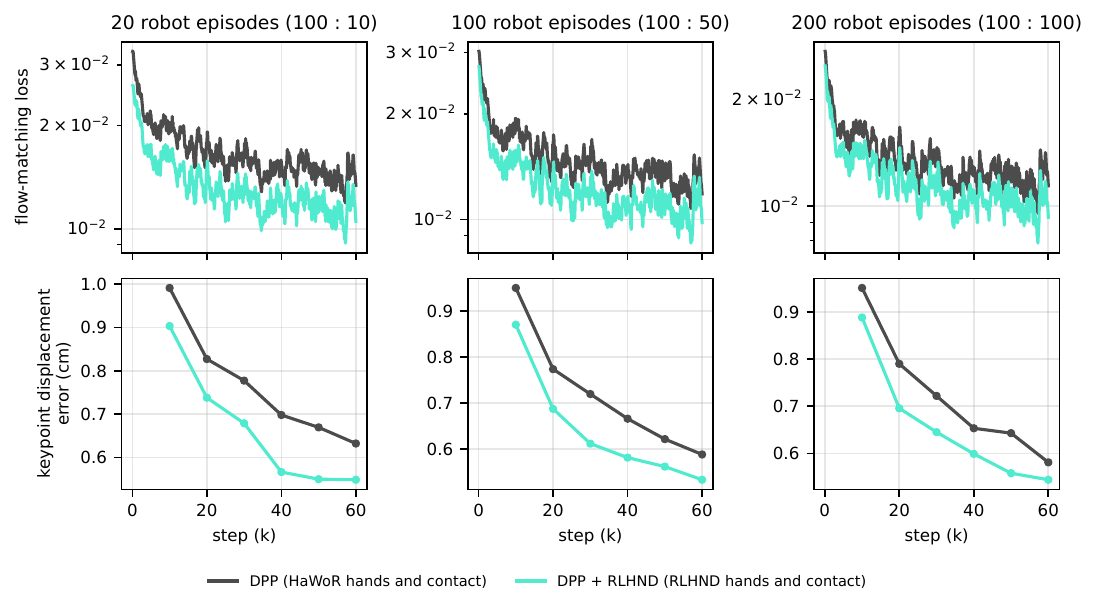}
\caption{
\textbf{Fine-tuning of DPP and DPP + RLHND.}
\emph{Top}: action flow-matching loss over $60$k fine-tuning steps.
\emph{Bottom}: keypoint displacement error of sampled actions at every $10$k-step snapshot, measured as the mean L2 distance in cm between predicted and recorded $16$-step displacements of the $21$ keypoints.
Results are shown for three human : robot ratios; only the 100 : 100 policies are deployed in Table~\ref{tab:robot_learning}.
Gray denotes DPP with its original human labels and teal denotes DPP + RLHND.
}

\label{fig:dpp_train_curves}
\end{figure}

\begin{figure}[t]
\centering
\includegraphics[width=\linewidth]{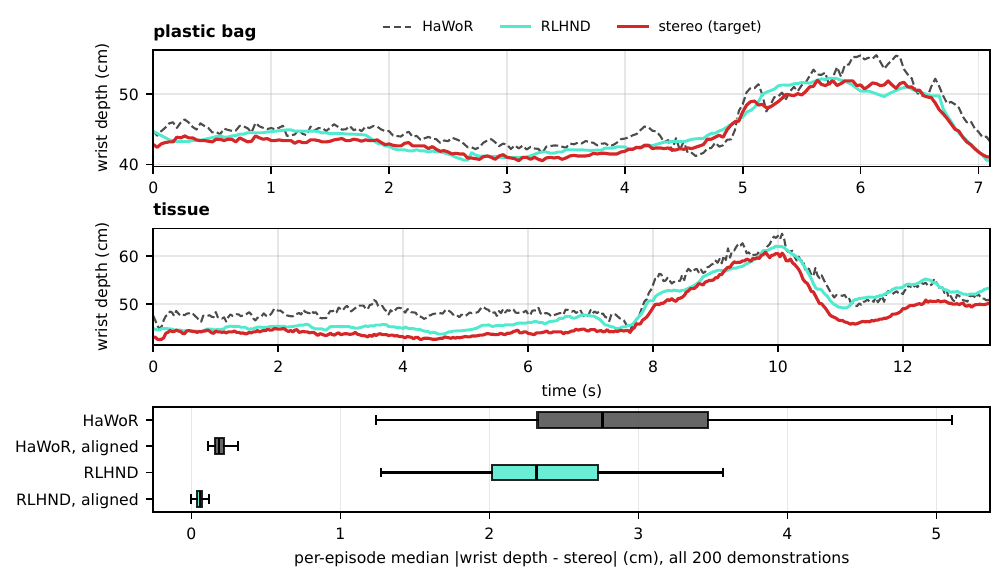}
\caption{
\textbf{Wrist depth of the human hand labels against the stereo depth.}
The top two rows show the right-wrist depth over one demonstration each of the plastic-bag and tissue tasks, for the \textcolor{stereored}{stereo target}, raw \textcolor{hawgray}{HaWoR}, and raw \textcolor{oursteal}{RLHND}, before the per-frame alignment.
The bottom row shows the per-episode median absolute deviation from the stereo target over all $200$ demonstrations, for both trackers before and after the alignment.
}
\label{fig:dpp_wrist_depth}
\end{figure}

\paragraph{Rollouts.}
Figure~\ref{fig:robot_demo} shows one rollout of the DPP + RLHND policy per task of Table~\ref{tab:robot_learning}. The pick-and-place rollouts share the same approach, grasp, lift, carry and release structure and differ only in the object; in the plastic-bag task the left hand picks up and flips the bag before handing it to the right hand, and in the tissue task the left hand removes the used core before the right hand places the new roll.

\paragraph{Depth of the human hand labels.}
Figure~\ref{fig:dpp_wrist_depth} compares the wrist depth of the two trackers with the stereo depth that the DPP pipeline aligns them to.
Even before the alignment, RLHND already follows the stereo trajectory closely.
Over the $200$ demonstrations, its right-wrist depth deviates from the stereo target by a median of $2.3$ cm, which is \emph{17.8\%} lower than HaWoR's.
In both cases the deviation is dominated by a constant offset, with a median bias of $+2.2$ cm for RLHND and $+2.6$ cm for HaWoR, whereas the frame-to-frame errors are visibly smaller for RLHND during fast reaching motions.
Since the alignment only rescales each frame and leaves the temporal noise of the tracker in place, the residual error is also smaller for RLHND after the per-frame alignment, \eg at $0.1$ cm against $0.2$ cm.
The remaining offset is largely a scale-depth ambiguity of the monocular hand, so it can be reduced further by measuring the actor's hand once and running RLHND with a pre-defined $\boldsymbol\beta_{\mathrm{cache}}$, which fixes the hand size and therefore the depth.
We nevertheless apply the same alignment to both sources, so that Table~\ref{tab:robot_learning} compares the trackers rather than their depth scale.

\paragraph{Fine-tuning loss.}
Figure~\ref{fig:dpp_train_curves} compares DPP fine-tuned with its original human labels and with RLHND labels, using both the action flow-matching loss and the keypoint displacement error of sampled actions on training windows.
Only the 100 : 100 policies of Table~\ref{tab:robot_learning} were deployed on the robot, but we fine-tuned both label sources for three human : robot ratios (100 : 10, 100 : 50, and 100 : 100, \ie $20$, $100$, and $200$ robot episodes).
DPP + RLHND consistently achieves a lower flow-matching loss across all ratios, and its sampled keypoint displacements remain closer to the recorded trajectories throughout training, reaching $0.55$, $0.53$, and $0.54$ cm versus $0.63$, $0.59$, and $0.58$ cm for DPP at 100 : 10, 100 : 50, and 100 : 100, respectively.
The deployed DPP + RLHND policy also achieves a higher rollout success rate in Table~\ref{tab:robot_learning}.

\subsection{Effect of the \texorpdfstring{$\boldsymbol\beta$}{beta}-Cache on Jitter}
\label{app:jitter_beta}
\begin{table}[t]
\centering
\setlength{\tabcolsep}{4pt}
\resizebox{\columnwidth}{!}{%
\begin{tabular}{lccccccccc}
\toprule
& \multicolumn{3}{c}{HOT3D} & \multicolumn{3}{c}{ARCTIC ego} & \multicolumn{3}{c}{EgoDex} \\
\cmidrule(lr){2-4}\cmidrule(lr){5-7}\cmidrule(lr){8-10}
Method & Jitter & Jitter$_z$ & Jitter$_{xy}$ & Jitter & Jitter$_z$ & Jitter$_{xy}$ & Jitter & Jitter$_z$ & Jitter$_{xy}$ \\
\midrule
HaMeR & 25.40 & 15.84 & 17.51 & 21.73 & 18.21 & 9.56 & 15.48 & 11.88 & 8.53 \\
WiLoR & 22.17 & 13.45 & 15.50 & 25.92 & 22.84 & 9.53 & 13.05 & 9.71 & \second{7.43} \\
HaWoR & 26.13 & 16.96 & 17.42 & 29.34 & 26.12 & 10.56 & 16.94 & 13.83 & 8.35 \\
HaPTIC & 38.27 & 27.02 & 23.97 & 51.09 & 47.28 & 16.40 & 25.81 & 21.88 & 12.06 \\
HandFlow & 16.99 & \second{8.41} & 13.37 & 13.47 & \second{7.88} & 9.49 & 10.89 & \second{5.69} & 8.32 \\
\midrule
\textbf{RLHND (Ours)} & \best{4.38} & \best{1.55} & \best{3.78} & \second{4.19} & \best{1.48} & \second{3.55} & \best{2.93} & \best{1.58} & \best{2.20} \\
\textbf{RLHND (Ours)} ($\boldsymbol\beta$-cache) & \second{4.39} & \best{1.55} & \second{3.79} & \best{4.18} & \best{1.48} & \best{3.53} & \second{2.94} & \best{1.58} & \best{2.20} \\
\bottomrule
\end{tabular}%
}
\caption{
\textbf{Jitter decomposition.} Jitter (mm/frame$^2$) is split into its depth component Jitter$_z$ and image-plane component Jitter$_{xy}$ under the protocol of Table~\ref{tab:pose_comparison}.
}
\label{tab:jitter_decomp}
\end{table}

Table~\ref{tab:pose_comparison} shows that the $\boldsymbol\beta$-cache removes within-recording shape drift, \eg $\sigma_{\mathrm{shape}}$ drops to zero, but leaves joint jitter essentially unchanged.
Table~\ref{tab:jitter_decomp} explains why by decomposing jitter into its depth component Jitter$_z$, the second temporal difference of the camera-frame $z$ coordinate, and its image-plane component Jitter$_{xy}$.

Three observations follow.
First, the residual jitter of clip-level trackers is dominated by the image plane.
For RLHND, Jitter$_{xy}$ accounts for at least three quarters of the total on every test set and is almost entirely a wrist-level effect. The wrist alone has $3.3$ / $3.2$ / $2.5$\,mm/frame$^2$ on HOT3D / ARCTIC / EgoDex, indicating that the 2D anchors driving mixed-PnP placement move slightly from frame to frame while the articulation remains stable.

Second, the $\boldsymbol\beta$-cache cannot affect either component by construction.
Depth is predicted directly by the per-frame depth head of the translation solver in Section~\ref{method:decoder} and does not depend on $\boldsymbol\beta$.
The cached shape only rescales the canonical joints, shifting the solved translation by a constant, sub-millimeter amount.
On HOT3D, the two RLHND rows differ by a median wrist displacement of $0.2$\,mm in $z$ and $0.4$\,mm in $xy$.
Because a constant shift within a recording has zero second temporal difference, Jitter$_z$ and Jitter$_{xy}$ coincide between the two rows to two decimals.
The benefit of the cache is therefore confined to size consistency measured by $\sigma_{\mathrm{shape}}$, while temporal stability comes from the clip-level decoding shared by both rows.

Third, the decomposition sharpens the comparison with crop-based trackers.
Their jitter is dominated by depth, reaching $18$-$47$\,mm/frame$^2$ on ARCTIC versus $1.5$ for RLHND, a $12$-$32\times$ gap, whereas their image-plane jitter is only $2$-$5\times$ higher.
Per-frame crops therefore stabilize apparent hand size but re-estimate metric distance independently at every frame, which directly affects the robot action space.

\subsection{Inference Cost}
\label{app:latency}
\begin{table}[t]
\centering
\small
\resizebox{\columnwidth}{!}{%
\begin{tabular}{lrrrrrr}
\toprule
Method & Params (M) & Detector (ms) & Model (ms) & Total (ms/frame) & FPS & MPJPE-p (HOT3D) \\
\midrule
HaMeR & 699 & 2.7 & 11.7 & 14.4 & 69.4 & 65.53 \\
WiLoR & 668 & 2.7 & 11.9 & 14.6 & 68.4 & 44.89 \\
HaWoR & 720 & 2.7 & 12.1 & 14.8 & 67.7 & 41.53 \\
HaPTIC & 1,357 & 2.7 & 23.8 & 26.5 & 37.7 & 62.42 \\
HandFlow & 866 & -- & 17.4 & 17.4 & 57.4 & 39.90 \\
ACE-Ego-Hand & 3,044 & -- & 1.8 & 1.8 & 568.6 & 24.41 \\
\midrule
\textbf{RLHND (Ours)} & 7,849 & -- & 6.9 & 6.9 & 144.5 & 13.01 \\
\bottomrule
\end{tabular}%
}
\caption{
\textbf{Inference cost.}
Inference time measured on one NVIDIA H100 (80\,GB) over the same HOT3D Aria test windows, using GPU-synchronized model-forward time per video frame with both hands.
Video decoding is excluded and the first forward pass is discarded as warm-up.
For crop-based trackers, the shared hand-detector cost is reported separately and included in the total.
MPJPE-p is from Table~\ref{tab:pose_comparison}.}
\label{tab:latency}
\end{table}

\begin{figure}[t]
\centering
\includegraphics[width=0.8\columnwidth]{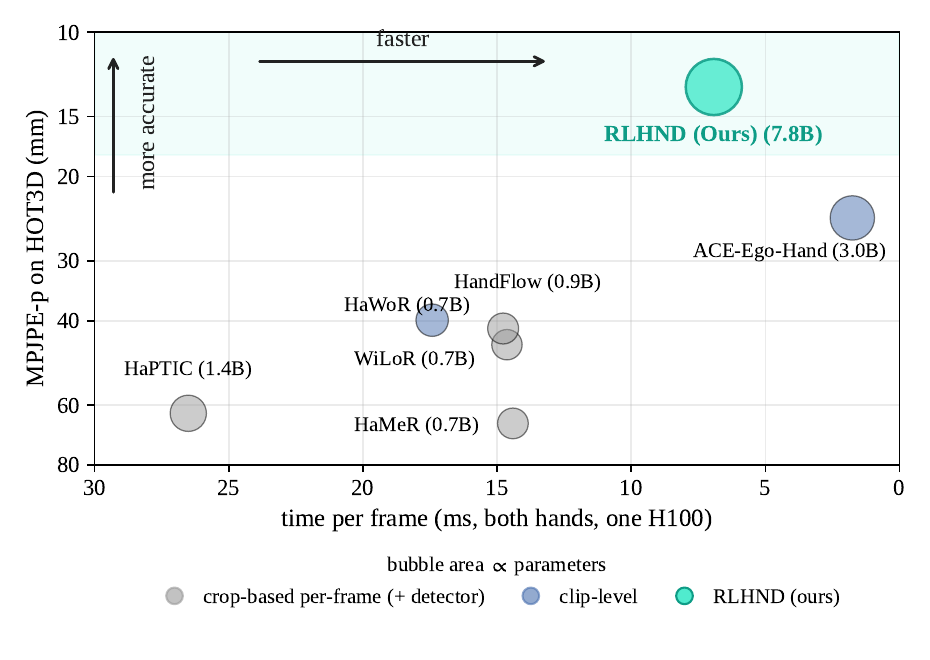}
\caption{
\textbf{Throughput against accuracy.}
Time per frame with both hands on one H100 (ms, linear, axis reversed so that faster is to the right) against MPJPE-p on HOT3D (log scale, lower is better, so better trackers lie further from the origin);
bubble area is proportional to the number of parameters used at inference (labeled in billions). Crop-based per-frame trackers include the shared hand detector.
Numbers in Table~\ref{tab:latency}.}
\label{fig:latency}
\end{figure}

Table~\ref{tab:latency} and Figure~\ref{fig:latency} report the inference cost of every tracker in Table~\ref{tab:pose_comparison} under a common protocol:
the same H100, the same HOT3D Aria test windows, GPU-synchronized model-forward time per video frame with both hands, and, for crop-based trackers, the shared hand detector included in the measurement.
RLHND runs its 7.8B-parameter Cosmos~3 Nano encoder once per 81-frame clip, so despite being the largest model, it has a per-frame cost comparable to a single ViT-H hand crop and lower cost than the crop-based trackers once the second hand and detector are included.
ACE-Ego-Hand, whose Wan2.2 encoder is truncated at a shallower depth, is the cheapest clip-level tracker at $4\times$ our throughput, but its HOT3D error is nearly twice ours.

\subsection{Per-Dataset Retargeting Results}
\label{app:retargeting_full}
\begin{table}[t]
\centering
\setlength{\tabcolsep}{3pt}
\resizebox{\columnwidth}{!}{%
\begin{tabular}{llcccccccccc}
\toprule
& & \multicolumn{2}{c}{Sharpa Wave (22 DoF)} & \multicolumn{2}{c}{WUJI v2 (20 DoF)} & \multicolumn{2}{c}{Shadow (24 DoF)} & \multicolumn{2}{c}{Inspire RH56 (12 DoF)} & \multicolumn{2}{c}{ALLEX (15 DoF)} \\
\cmidrule(lr){3-4}\cmidrule(lr){5-6}\cmidrule(lr){7-8}\cmidrule(lr){9-10}\cmidrule(lr){11-12}
Test set & Method & Q-err $\downarrow$ & Jerk $\downarrow$ & Q-err $\downarrow$ & Jerk $\downarrow$ & Q-err $\downarrow$ & Jerk $\downarrow$ & Q-err $\downarrow$ & Jerk $\downarrow$ & Q-err $\downarrow$ & Jerk $\downarrow$ \\
\midrule
\multirow{9}{*}{HOT3D}
 & \textcolor{gray}{Oracle (GT joints)} & \textcolor{gray}{0.0} & \textcolor{gray}{0.27} & \textcolor{gray}{0.0} & \textcolor{gray}{0.22} & \textcolor{gray}{0.0} & \textcolor{gray}{0.27} & \textcolor{gray}{0.0} & \textcolor{gray}{0.13} & \textcolor{gray}{0.0} & \textcolor{gray}{0.25} \\
 & HaMeR~\citep{pavlakos2024reconstructing} & 15.0 & 1.50 & 17.8 & 1.46 & 13.1 & 1.30 & 7.1 & 0.99 & 13.3 & 1.36 \\
 & WiLoR~\citep{potamias2024wilor} & 11.8 & 1.25 & 14.6 & 1.24 & 9.4 & 1.03 & 4.9 & 1.00 & 9.6 & 1.11 \\
 & HaWoR~\citep{zhang2025hawor} & 13.6 & 0.37 & 15.8 & 0.32 & 10.7 & 0.36 & 7.2 & 0.25 & 11.2 & 0.32 \\
 & HaPTIC~\citep{ye2025haptic} & 15.3 & 1.60 & 18.2 & 1.52 & 12.7 & 1.09 & 6.8 & 1.10 & 13.7 & 1.66 \\
 & HandFlow~\citep{xu2026handflow} & 14.8 & 0.56 & 16.5 & 0.54 & 14.5 & 0.46 & 10.4 & 0.38 & 15.5 & 0.59 \\
 & ACE-Ego-Hand~\citep{liu2026aceegohand} & 10.0 & \second{0.24} & 12.4 & 0.25 & \second{8.5} & 0.26 & 4.9 & 0.12 & 8.8 & 0.23 \\
\cmidrule(lr){2-12}
 & \textbf{RLHND (Ours)} & \best{8.9} & \best{0.22} & \best{10.2} & \best{0.17} & \best{7.4} & \best{0.17} & \second{4.0} & \second{0.08} & \best{6.9} & \second{0.19} \\
 & \textbf{RLHND (Ours)} ($\boldsymbol\beta$-cache) & \second{9.0} & \best{0.22} & \second{11.2} & \second{0.18} & \best{7.4} & \second{0.18} & \best{3.9} & \best{0.07} & \second{7.1} & \best{0.17} \\
\midrule
\multirow{9}{*}{ARCTIC ego}
 & \textcolor{gray}{Oracle (GT joints)} & \textcolor{gray}{0.0} & \textcolor{gray}{0.34} & \textcolor{gray}{0.0} & \textcolor{gray}{0.38} & \textcolor{gray}{0.0} & \textcolor{gray}{0.36} & \textcolor{gray}{0.0} & \textcolor{gray}{0.23} & \textcolor{gray}{0.0} & \textcolor{gray}{0.44} \\
 & HaMeR~\citep{pavlakos2024reconstructing} & 14.0 & 1.59 & 17.0 & 1.43 & 11.6 & 1.50 & 8.6 & 1.25 & 14.1 & 1.56 \\
 & WiLoR~\citep{potamias2024wilor} & 10.7 & 1.57 & \second{14.1} & 1.40 & \second{9.7} & 1.12 & \second{6.4} & 1.00 & 10.3 & 1.49 \\
 & HaWoR~\citep{zhang2025hawor} & 13.2 & 0.95 & 16.8 & 0.74 & 11.8 & 0.72 & 11.0 & 0.71 & 14.3 & 0.77 \\
 & HaPTIC~\citep{ye2025haptic} & \second{10.2} & 1.47 & \best{13.2} & 1.37 & 10.3 & 1.36 & \best{6.2} & 1.20 & \second{10.1} & 1.32 \\
 & HandFlow~\citep{xu2026handflow} & 23.1 & 0.57 & 26.9 & 0.63 & 16.3 & 0.59 & 12.6 & 0.81 & 22.1 & 0.63 \\
 & ACE-Ego-Hand~\citep{liu2026aceegohand} & 12.0 & 0.21 & 17.2 & 0.24 & 12.1 & \best{0.21} & 9.1 & 0.16 & 12.4 & \second{0.25} \\
\cmidrule(lr){2-12}
 & \textbf{RLHND (Ours)} & \best{9.3} & \second{0.18} & 15.4 & \best{0.18} & \second{9.7} & \second{0.23} & 6.8 & \best{0.08} & \best{9.4} & \best{0.20} \\
 & \textbf{RLHND (Ours)} ($\boldsymbol\beta$-cache) & \best{9.3} & \best{0.15} & 15.2 & \second{0.20} & \best{9.6} & \second{0.23} & 6.8 & \second{0.10} & \best{9.4} & \best{0.20} \\
\midrule
\multirow{9}{*}{EgoDex}
 & \textcolor{gray}{Oracle (GT joints)} & \textcolor{gray}{0.0} & \textcolor{gray}{0.42} & \textcolor{gray}{0.0} & \textcolor{gray}{0.40} & \textcolor{gray}{0.0} & \textcolor{gray}{0.43} & \textcolor{gray}{0.0} & \textcolor{gray}{0.17} & \textcolor{gray}{0.0} & \textcolor{gray}{0.39} \\
 & HaMeR~\citep{pavlakos2024reconstructing} & 21.0 & 1.27 & 23.5 & 1.05 & 13.2 & 1.14 & 8.5 & 0.47 & 15.0 & 0.66 \\
 & WiLoR~\citep{potamias2024wilor} & 20.0 & 1.62 & 22.2 & 1.37 & \best{11.6} & 1.46 & 7.7 & 0.71 & \best{13.8} & 0.95 \\
 & HaWoR~\citep{zhang2025hawor} & \best{18.3} & 0.52 & \second{21.6} & \second{0.35} & 14.2 & 0.42 & 8.6 & 0.22 & 17.5 & 0.40 \\
 & HaPTIC~\citep{ye2025haptic} & \second{19.2} & 1.10 & \best{21.3} & 0.91 & 12.5 & 0.85 & 8.1 & 0.31 & \second{14.3} & 0.66 \\
 & HandFlow~\citep{xu2026handflow} & \best{18.3} & 0.62 & 22.0 & 0.48 & 11.8 & 0.49 & \best{6.8} & 0.29 & 15.8 & 0.38 \\
 & ACE-Ego-Hand~\citep{liu2026aceegohand} & 20.1 & \second{0.15} & 22.4 & \best{0.14} & \second{11.7} & \best{0.16} & 8.4 & 0.05 & 14.8 & \best{0.12} \\
\cmidrule(lr){2-12}
 & \textbf{RLHND (Ours)} & 20.0 & \best{0.14} & 22.8 & \best{0.14} & \best{11.6} & 0.18 & \second{7.4} & \best{0.03} & 14.4 & \second{0.13} \\
 & \textbf{RLHND (Ours)} ($\boldsymbol\beta$-cache) & 20.0 & \second{0.15} & 22.6 & \best{0.14} & \second{11.7} & \second{0.17} & \second{7.4} & \second{0.04} & 14.5 & 0.14 \\
\bottomrule
\end{tabular}%
}
\caption{Per-dataset retargeting quality on five dexterous right hands (DexPilot, wrist-relative targets); Table~\ref{tab:retargeting} in the main text reports the mean of the three blocks. Q-err is the mean absolute deviation ($^\circ$) of the joint command from the one the same solver produces on the ground-truth joints; Jerk is the median over frames of the joint-averaged second temporal difference of the command ($^\circ$/frame$^2$), i.e.\ the frame-to-frame noise on typical frames (Appendix~\ref{app:metrics}). The grey Oracle row retargets the ground-truth joints (Q-err $=0$ by construction). We denote \bestcap{best} and \secondcap{second best} values with shade and bold.}
\label{tab:retargeting_full}
\end{table}

Table~\ref{tab:retargeting_full} breaks down the retargeting comparison of Table~\ref{tab:retargeting} by test set, \ie HOT3D, ARCTIC ego, and EgoDex, using the same DexPilot solver, wrist-relative targets, and metrics as in the main text.
On HOT3D, RLHND has the lowest or second-lowest Q-err on every embodiment; on ARCTIC ego it has the lowest Q-err on Sharpa, Shadow, and ALLEX, while HaPTIC and WiLoR are closer on WUJI and Inspire. On both sets RLHND, together with ACE-Ego-Hand, produces the smoothest commands.
On EgoDex, Q-err differences are smaller, with all trackers within $3^\circ$ of each other on WUJI and Sharpa, likely due to noise inherent in the ARKit ground-truth joints themselves.
The Oracle row has jerk of $0.13$-$0.27$\,deg/frame$^2$ on HOT3D and $0.17$-$0.43$\,deg/frame$^2$ on EgoDex, showing that the reference commands used for Q-err already contain frame-to-frame noise.

RLHND remains within $2^\circ$ of the best method on every embodiment on EgoDex and achieves the lowest Q-err on Shadow.
It also ties ACE-Ego-Hand for the lowest jerk, with each method achieving the lowest jerk on some of the five hands.
Across all three test sets, the crop-based per-frame trackers exhibit substantially higher jerk than the clip-level trackers HandFlow, ACE-Ego-Hand, and RLHND, with differences of the same order of magnitude.

\subsection{Additional Qualitative Results}
\label{app:qualitative}
\begin{figure}[p]
\centering
\includegraphics[width=0.97\linewidth]{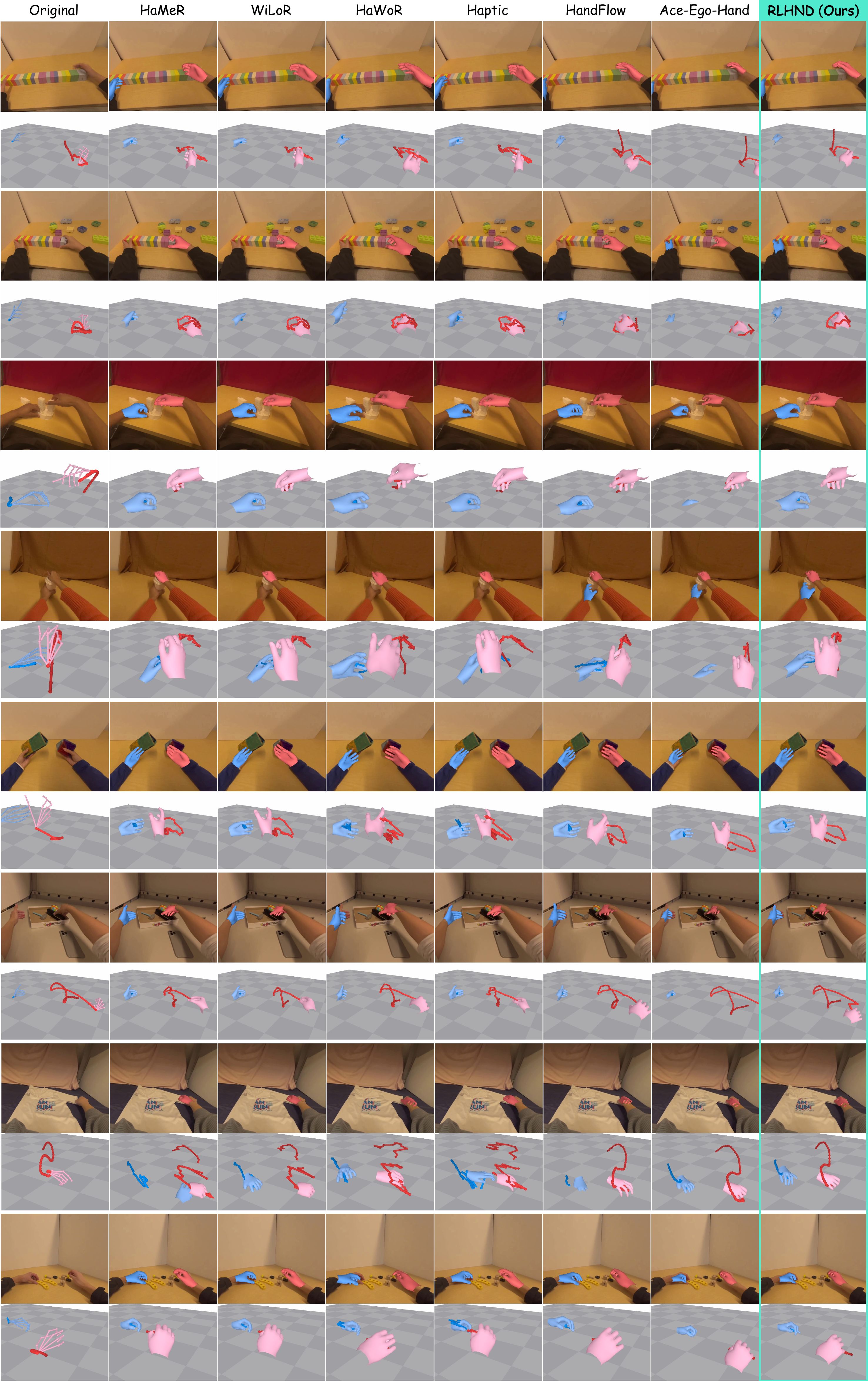}
\caption{
\textbf{Extensive qualitative comparison of 2D and 3D hand motion reconstruction.}
}
\label{fig:qualitative_pose_app_01}
\end{figure}

Figure~\ref{fig:qualitative_pose_app_01} extends Figure~\ref{fig:qualitative_pose} with additional EgoDex windows from a test set unseen during training.
The mesh overlays are similar across methods, as crop-based trackers are optimized primarily for per-frame reprojection.
Their trajectories, however, show clearer differences.
The crop-based trackers, \ie HaMeR, WiLoR, HaWoR, and HaPTIC, estimate each frame independently and exhibit jagged wrist trajectories, consistent with the depth instability quantified in Figure~\ref{fig:motivation} and reflected in the jerk metric of Table~\ref{tab:retargeting_full}.
ACE-Ego-Hand produces smoother trajectories but misses a hand in several windows, resulting in missing or displaced meshes, whereas RLHND tracks both hands more consistently and more closely follows the ground-truth trajectories.

\begin{figure}[p]
\centering
\includegraphics[width=0.98\linewidth]{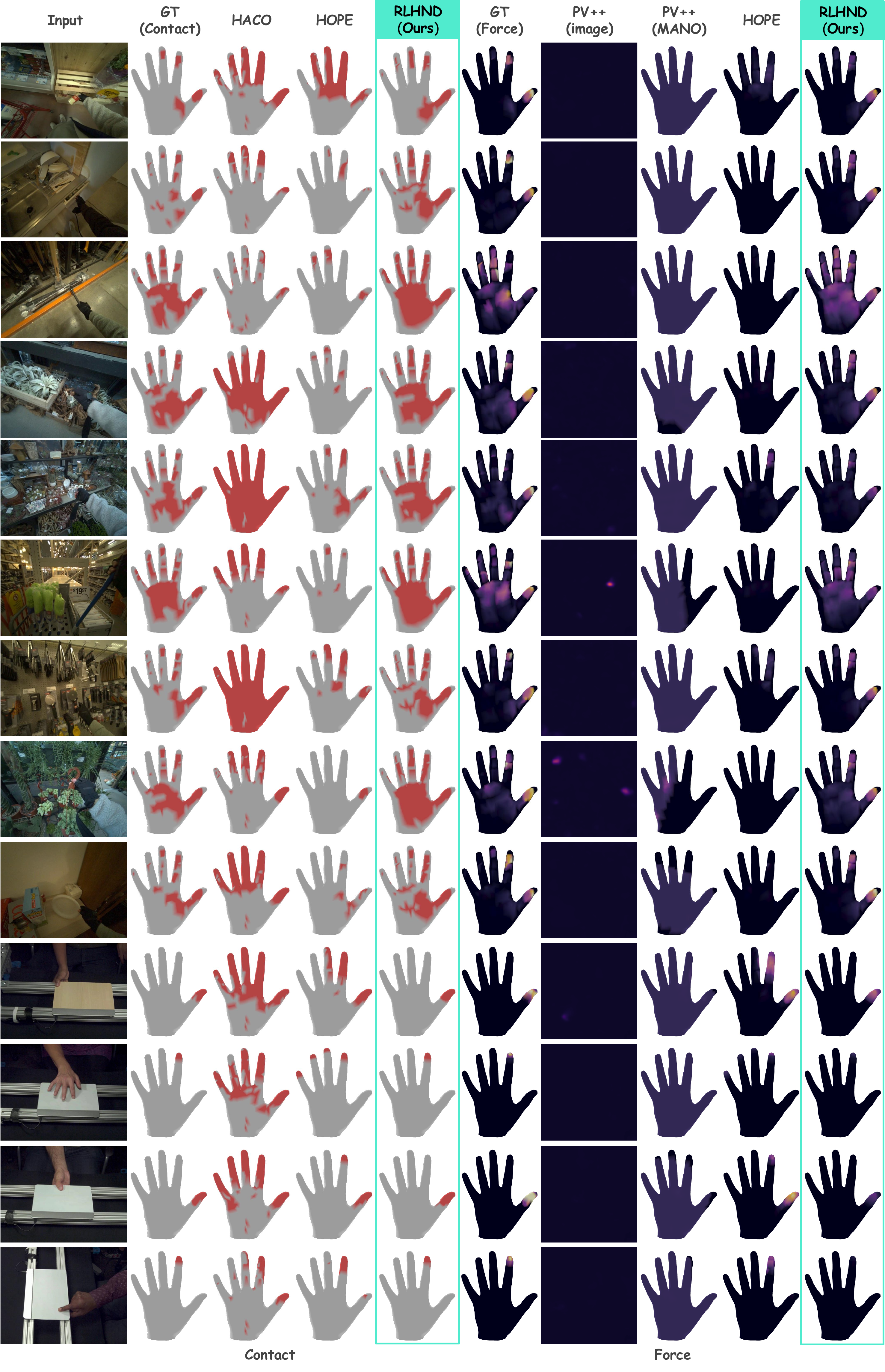}
\vspace{-0.3cm}
\caption{
\textbf{Extensive qualitative comparison of per-vertex contact and force estimation.}
}
\label{fig:qualitative_tactile_app_01}
\end{figure}

Figure~\ref{fig:qualitative_tactile_app_01} extends Figure~\ref{fig:qualitative_tactile} with additional OpenTouch and PVDB test frames.
The same qualitative differences appear across scenes: HACO predicts contact over broader regions of the palm and fingers, HOPE produces fragmented contact regions with lower pressure estimates, and PressureVision++ produces little output when the hand is outside its planar-sensor representation.
In contrast, RLHND places contact and pressure in regions corresponding to the glove or pressure pad measurements, including the single-fingertip presses in the PVDB examples.

\subsection{Additional Ablation Results}
\label{app:ablation_full}
\paragraph{Per-vertex image sampling.}
We additionally evaluate per-vertex image features to test whether local visual evidence improves tactile estimation.
We augment Eq.~(\ref{eq:lbs_spread}) with a zero-initialized MLP applied to decoder features bilinearly sampled at each vertex's projected image location, together with a visibility cue given by the cosine between the vertex normal and the viewing ray.
This variant achieves vertex-level contact F1 of $0.709$ / $0.543$ / $0.570$ on OpenTouch / DexYCB / HOT3D and an OpenTouch force MAE of $1.256$\,kPa, compared with $0.696$ / $0.572$ / $0.589$ and $0.489$\,kPa for the default readout in Table~\ref{tab:ablation}.
Replacing the nearest-latent lookup with temporal interpolation or removing the visibility cue changes every metric by less than $0.01$.
These results suggest that per-vertex image evidence provides little additional information beyond the pose-conditioned features and label supervision, so we omit the image-sampling branch from the final design.

\paragraph{Conditioning the expert on the pose stream.}
HOPE~\citep{jeon2026hope} ablates the inputs of its vertex transformer and reports that combining visual features with hand pose outperforms either alone, suggesting that the two provide complementary information for contact and pressure estimation.
Our expert already receives pose information indirectly, as its bone tokens are initialized from the pose decoder and spread to vertices through the MANO skinning weights.
We therefore test whether directly attending to the pose tokens provides additional information.
We implement this as \emph{one-way} attention, where the expert's bone tokens take the pose tokens as additional keys and values, while the pose tokens never attend back.
Thus, the frozen stage-1 stream remains unchanged by the tactile objective and the pose predictions are bit-identical.
Table~\ref{tab:oneway} compares the two variants using the same checkpoint, data, and training recipe.

The one-way variant performs worse on OpenTouch ($0.635$ vs.\ $0.696$ F1) and DexYCB ($0.565$ vs.\ $0.572$), improves by $0.004$ on HOT3D, and has higher force MAE on OpenTouch ($0.573$ vs.\ $0.489$\,kPa).
AUROC is nearly unchanged across all three splits ($0.979$/$0.910$/$0.955$ vs.\ $0.980$/$0.915$/$0.959$), indicating that the additional pathway does not materially change the ranking of vertices and mainly shifts the operating point without a consistent direction.
These results suggest that the pose information available through the existing bone tokens is already sufficient, and that directly attending to the pose tokens provides little additional signal.
We therefore omit the additional attention pathway from the final expert.

\begin{table}[h]
\centering
\setlength{\tabcolsep}{4pt}
\resizebox{\columnwidth}{!}{%
\begin{tabular}{lcccccccc}
\toprule
& \multicolumn{2}{c}{OpenTouch} & \multicolumn{2}{c}{DexYCB} & \multicolumn{2}{c}{HOT3D} & \multicolumn{2}{c}{OpenTouch force (kPa)} \\
\cmidrule(lr){2-3}\cmidrule(lr){4-5}\cmidrule(lr){6-7}\cmidrule(lr){8-9}
Variant & F1 $\uparrow$ & AUROC $\uparrow$ & F1 $\uparrow$ & AUROC $\uparrow$ & F1 $\uparrow$ & AUROC $\uparrow$ & MAE $\downarrow$ & RMSE $\downarrow$ \\
\midrule
w/ one-way attention & 0.635 & 0.979 & 0.565 & 0.910 & \best{0.593} & 0.955 & 0.573 & 2.574 \\
\textbf{Ours} (no pose conditioning) & \best{0.696} & \best{0.980} & \best{0.572} & \best{0.915} & 0.589 & \best{0.959} & \best{0.489} & \best{2.508} \\
\bottomrule
\end{tabular}%
}
\caption{
\textbf{One-way attention ablation.}
Comparison between the proposed tactile expert and a variant in which the bone tokens additionally attend to the pose stream.
\bestcap{Best} values are shaded.
}

\label{tab:oneway}
\end{table}

\section{Limitations}
\label{app:limitations}
\paragraph{Shape calibration on public benchmarks.}
The $\boldsymbol\beta$-cache is designed around a hand shape measured once per operator, which is natural in robot data collection but unavailable on public benchmarks without per-subject calibration. We therefore initialize the cache from the first well-visible window on these benchmarks, and can demonstrate the benefit of external calibration only using the ground-truth $\boldsymbol\beta$ available in ARCTIC (A4 in Table~\ref{tab:ablation}). The benefit of external calibration in unconstrained settings, and the robustness of the cache when the first well-visible window provides a poor view of the hand, remain to be evaluated.

\paragraph{Scarce tactile supervision.}
Force supervision in our training data comes from a single tactile glove (OpenTouch) and a single planar pressure pad (PressureVisionDB), limiting the diversity of pressure magnitudes and contact surfaces seen during training. Newer datasets such as EgoPressure~\citep{zhao2025egopressure} provide additional egocentric pressure data, but incorporating them requires reconciling their sensor calibration, surface parameterization, and contact definitions with those of existing datasets. Until tactile datasets adopt more consistent surface representations and pressure units, scaling force supervision across heterogeneous sensors remains challenging.

\paragraph{From hand labels to robot actions.}
Accurate hand motion and contact labels do not directly determine robot actions. Our real-robot experiments use a fixed inverse-kinematics retargeting procedure with locked abduction joints and a heuristic contact offset, while the retargeting results in Table~\ref{tab:retargeting} measure fidelity to a solver applied to the ground-truth hand rather than task success. Mapping human hand motion and contact patterns to dexterous hands with different kinematics, contact geometry, and compliance, as well as exploiting predicted force beyond binary contact, remains outside the scope of this work. Finally, RLHND processes clips offline rather than in real time, and its backbone is substantially larger than crop-based regressors.

\end{document}